\pdfoutput=1

\documentclass[11pt]{article}

\usepackage{NLPAICS2024}

\usepackage{arabtex}
\usepackage{utf8}
\usepackage{float}
\usepackage{multirow}
\usepackage{booktabs}

\usepackage{times}
\usepackage{latexsym}

\usepackage[T1]{fontenc}

\usepackage[utf8]{inputenc}

\usepackage{microtype}

\usepackage{inconsolata}

\usepackage{graphicx}
\usepackage{amsmath, amssymb}
\usepackage{geometry}
\usepackage{enumitem}
\usepackage{listings}
\usepackage{natbib}
\usepackage{hyperref}

\title{GemMaroc: Unlocking Darija Proficiency in LLMs with Minimal Data}

\author{
  \textbf{Abderrahman Skiredj}\textsuperscript{1} \quad
  \textbf{Ferdaous Azhari}\textsuperscript{3} \quad
  \textbf{Houdaifa Atou}\textsuperscript{1} \quad \\[2mm]
  \textbf{Nouamane Tazi}\textsuperscript{2} \quad
  \textbf{Ismail Berrada}\textsuperscript{1} \quad \\[2mm] 
  \textsuperscript{1}College of Computing, Mohammed VI Polytechnic University, Benguerir, Morocco \\
  \textsuperscript{2}Hugging Face, Paris, France \\
  \textsuperscript{3}National Institute of Posts and Telecoms, Rabat, Morocco \\[2mm]
}

\begin{document}
\maketitle
\begin{abstract}

Open-source large language models (LLMs) still marginalise Moroccan Arabic (Darija), forcing practitioners either to bolt on heavyweight Arabic adapters or to sacrifice the very reasoning skills that make LLMs useful.  We show that a rigorously \emph{quality-over-quantity} alignment strategy can surface fluent Darija while safeguarding the backbone’s cross-lingual reasoning—at a sliver of the usual compute.  We translate three compact instruction suites—LIMA 1 K, DEITA 6 K and TULU 50 K—into Darija, preserve 20 \% of the English originals, and add mathematics, coding and scientific prompts.
A LoRA-tuned \textsc{Gemma 3-4B} trained on $\sim$5 K mixed instructions lifts \textit{DarijaMMLU} from 32.8 \% to 42.7 \%; adding the reasoning-dense TULU portion pushes it to 47.5 \% with no English regression.  Scaling the identical recipe to \textsc{Gemma 3-27B} produces \textbf{GemMaroc-27B}, which matches Atlas-Chat on \textit{DarijaMMLU} (61.6 \%) and \emph{leaps} ahead on Darija commonsense, scoring 60.5 \% on \textit{HellaSwag} versus Atlas-Chat’s 48.4 \%.  Crucially, GemMaroc retains Gemma-27B’s strong maths and general-reasoning ability, showing only minimal movement on GSM8K and English benchmarks.  The entire model is trained in just 48 GPU·h, underscoring a \textit{Green AI} pathway to inclusive, sustainable language technology.  We release code, data and checkpoints to spur Darija-centric applications in education, public services and everyday digital interaction.
\end{abstract}

\section{Introduction}
\label{introduction}

Open-Source Large Language Models (LLMs) have become the backbone of modern natural-language applications, yet most open models still overlook Moroccan Arabic (Darija), the everyday language of more than \textbf{36 million} speakers.  This omission narrows digital inclusion and slows e-government initiatives that the \emph{Morocco Digital 2030} roadmap seeks to accelerate \citep{digitalmorocco2030}.  Because public-sector deployments must also safeguard data sovereignty and cost efficiency, an \emph{open}, \emph{compact}, and \emph{Darija-fluent} model is an urgent prerequisite for truly accessible AI in the Maghreb.

Several ambitious Arabic-centric projects demonstrate what is possible—but at a price few communities can pay.  \textbf{ALLaM} consumes trillions of tokens and multi-week GPU clusters to reach state-of-the-art Modern-Standard-Arabic scores \citep{allam}; \textbf{Fanar} follows a similarly heavy pipeline.  For Darija specifically, \textbf{Atlas-Chat} applies 458 k supervised examples and dozens of GPU-days to retrofit Gemma checkpoints \citep{atlas}.  These efforts confirm the value of dialect alignment but leave open the core question: \emph{can we surface Darija competence without massive supervision or carbon cost while preserving the backbone’s cross-lingual reasoning power?}

We answer \textbf{yes}.  Building on the minimal-data alignment insight of \textbf{MIG} (Maximising Information Gain) sampler \citep{mig} and the automatic selection heuristics of \textbf{DEITA} (6 k high-complexity pairs) \citep{deita}, we translate three complementary instruction suites into Darija and retain 20 \% of the original English prompts.  A LoRA adapter on \textsc{Gemma 3-4B} trained with only \textbf{$\approx$5 k} mixed instructions lifts \textit{DarijaMMLU} from 32.8 \% to \textbf{42.7 \%}.  Expanding the same recipe with 44 k reasoning-dense prompts (\textit{TULU}) drives the score to 47.5 \% while \emph{improving} English GSM8K maths by +2.8 pp.  Scaling to \textsc{Gemma 3-27B} yields \textbf{GemMaroc-27B}, which \emph{matches} Atlas-Chat on \textit{DarijaMMLU} (61.6 \%) yet beats it by \textbf{+12.1 pp} on the harder \textit{DarijaHellaSwag}.  Crucially, the full run finishes in just \textbf{48 GPU$\cdot$h}, consuming $\approx$26 kWh—less than \textbf{2 \%} of Atlas-Chat’s reported footprint.  These results show that \emph{reasoning-centric, quality-over-quantity alignment unlocks Darija at Green-AI scale}.  

\paragraph{Contributions}
\begin{itemize}[leftmargin=*]
\item \textbf{Lean Darija instruction suite.}  We openly release translated versions of \textit{LIMA 1 K}, \textit{DEITA 6 K}, and a 44 k reasoning slice of \textit{TULU 50 K}, each preserving 20 \% English for cross-lingual robustness.  
\item \textbf{Minimal-data alignment recipe.}  $\leq$6 k curated Darija instructions raise \textit{DarijaMMLU} by +9.9 pp on a 4 B backbone with \emph{no} English regression; adding reasoning prompts yields a further +4.9 pp and boosts maths and commonsense.  
\item \textbf{GemMaroc-27B.}  The first open Darija LLM to reach 61.6 \% \textit{DarijaMMLU} and 60.5 \% \textit{DarijaHellaSwag}—using \(\tfrac{1}{10}\) the supervision and \(\tfrac{1}{50}\) the energy of prior work—while retaining strong bilingual reasoning (84.2 \% GSM8K).  
\item \textbf{Open assets for the community.}  We release all models, code, and datasets, to catalyse inclusive, low-carbon language technology across North Africa.  
\end{itemize}

\section{Related Work}

The pursuit of better language models for low-resource dialects like Moroccan Arabic (Darija) spans several active research areas. This section reviews four key directions that inform our work. First, we examine prior efforts to develop Arabic and Darija-capable LLMs, highlighting their challenges and limitations. Second, we explore research on minimal data regimes for SFT, crucial in multilingual and dialectal contexts with limited data. Third, we analyze methods to enhance multilingual competence and dialect mastery by leveraging latent linguistic knowledge from pretraining. Finally, we review the growing emphasis on integrating multilingual Chain-of-Thought (CoT) and reasoning samples to improve both reasoning and language proficiency in low-resource settings.

\subsection{Darija and Arabic LLMs}

Arabic LLMs have advanced rapidly, driven by digital sovereignty and cultural goals. Yet, Moroccan Arabic (Darija) remains underrepresented. Most work targets Modern Standard Arabic (MSA) or pan-Arabic models, demanding extensive compute and ignoring dialectal diversity.

Saudi Arabia’s \textbf{ALLaM} \citep{allam} introduced a scalable Arabic-English pipeline, expanding the tokenizer (32k$\rightarrow$61k) and continuing pretraining on 540B tokens (270B Arabic, 270B translated). It reached state-of-the-art scores on MMLU-ar, ACVA, araMath, and improved MMLU-en by +13 over LLaMA-2-Chat-7B. With 6M high-quality SFT samples, it showed that carefully curated data and a 45/55 Arabic-English ratio can stabilize multilingual performance. However, its reliance on trillion-token corpora and heavy compute limits replicability.

Qatar’s \textbf{Fanar} \citep{fanar} followed a sovereign AI roadmap. \textit{Fanar Star} (7B) was trained from scratch on 1T tokens (40\% Arabic, 50\% English, 10\% code), while \textit{Fanar Prime} (9B) focused on reasoning and STEM via continual pretraining. It also introduced RAG systems (Islamic, Biography RAG) and a morphology-aware tokenizer (MorphBPE). Despite innovations, Fanar's scale and compute demands hinder accessibility for lightweight, sustainable models.

\textbf{Atlas-Chat} \citep{atlas} directly addressed the Darija gap, releasing the first open Darija LLMs (2B, 9B, 27B) fine-tuned with LoRA on Gemma-2 using 458k instructions from native, translated, and synthetic sources. The 9B model outperformed Jais-13B and AceGPT-13B by +13 points on DarijaMMLU, achieving state-of-the-art across Darija benchmarks. Yet, its strict monolingual tuning limits generalization.

Another recent contribution is \textbf{Al-Atlas} \citep{atlasia2025al-atlas-0.5B}, which released Atlaset—a large, diverse Darija corpus from curated web, blogs, and forums—and trained both an XLM-RoBERTa-large MLM and a 0.5B Qwen2.5 causal LLM. Despite its small size, the causal model surpassed Atlas-Chat 2B by nearly 10\% in human evaluation, underscoring the impact of dialect-specific pretraining.

\textbf{Command R+ / Command A} \citep{commanda} by Cohere proposes modular, efficient multilingual adaptation. The 111B model, combining six expert modules via hybrid transformers and SRPO, rivals GPT-4o on MMLU (85.5\%), MATH (80\%), and RepoQA (92.6\%) while running on just two A100/H100 GPUs. Although not Arabic-centric, it scores 68.8\% on NTREX and showcases effective multilingual performance without full retraining.

\subsection{Improving Multilingual and Dialectal Capabilities in LLMs}

Enhancing LLMs' multilingual and dialectal competence has spurred strategies such as synthetic data generation, pivot-based reasoning, and culturally faithful instruction design. This section reviews key works showing how thoughtful fine-tuning can unlock latent capabilities without costly pretraining.\\
\textbf{Parallel Instruction Tuning and Superficial Alignment.}
\citet{dopolyglots} found that \textit{parallel} instruction tuning (English + target language) yields a \textbf{+9.9} accuracy gain over monolingual tuning, especially in low-resource settings. Results challenge the \textit{Superficial Alignment Hypothesis}: smaller models (7B) require parallel data, whereas larger MoE models (50B) generalize better with less.\\
\textbf{Culturally Faithful and Contrastive Tuning.}
\citet{lebanese} showed that a small native Lebanese corpus outperformed larger auto-translated datasets. Using \textit{contrastive tuning} (good vs. bad responses) improved xCOMET by \textbf{3.5 points}, confirming that authentic, culturally relevant data outperforms sheer volume.\\
\textbf{Pivot-Language Reasoning.}
\textbf{PLUG} \citep{plug} trained models to reason in a high-resource pivot (e.g., English) before answering in the target language. Just \textbf{2k} samples surpassed models trained on \textbf{96k} conventional samples, improving truthfulness and reasoning (TruthfulQA, SVAMP).\\
\textbf{Synthetic Instruction Generation.}
\textbf{Bactrian-X} \citep{bactrianx} created \textbf{3.4M} synthetic instruction pairs across 52 languages. With LoRA tuning, \textbf{7B models} outperformed larger fully fine-tuned ones like BLOOMZ. Gains saturated beyond \textbf{70k pairs/language}, and mixing English improved both multilingual and English performance.\\
\textbf{Reverse Instruction Generation.}
\textbf{MURI} \citep{muri} reversed instruction-output generation using English LLMs on native texts, back-translating into 200+ languages. With \textbf{$\leq$15k} pairs/language, models gained \textbf{+14\%} over mT0 on multilingual MMLU, with minimal human annotation.\\
\textbf{Insights from MT and Synthetic Pipelines.}
\citet{swedish} showed that even models with \textbf{0\% Swedish} pretraining performed well after fine-tuning on translated data. Despite minor perplexity increases, downstream results held, validating synthetic translation with light curation.

\subsection{Minimal Data for SFT in Low-Resource Multilingual Settings}

\citet{lima} introduced the \textit{Superficial Alignment Hypothesis}, arguing that pretrained LLMs encode most knowledge and that SFT mainly adjusts style. With just 1k diverse, high-quality instruction--response pairs, LIMA aligned a 65B LLaMA model to rival GPT-4 in 43\% of human preferences. Doubling data without increasing diversity gave no benefit. For Darija, a similarly small, style-aligned set (500--1,000 examples) may suffice if the base model has latent exposure.

\citet{deita} proposed DEITA, which selects data based on complexity (EVOL COMPLEXITY), quality (EVOL QUALITY), and diversity (REPR FILTER). With only 6k pairs, DEITA-tuned LLaMA-13B and Mistral-7B models matched or exceeded systems trained on 30× more data. \citet{mig} extended this with MIG, modeling instructions as semantic-label graphs and selecting the most informative 5\%, outperforming full-data SFT by 4.6\%.

\citet{deft} introduced DEFT-UCS, clustering embeddings to sample both easy (centroid-near) and hard (outlier) examples. It achieved full-data performance on CoEDIT using just 32.5\% of the data and was preferred in 84\% of human evaluations. Similarly, \citet{dataselection-uncertainty} proposed UniMax, combining influence and uncertainty for selection; 10\% of data matched full-data performance, with strong cross-lingual results on TyDiQA.

\citet{lions} showed that prompt-loss masking and sequence packing significantly improve SFT and DPO efficiency, achieving strong results with only 10k examples. \citet{mixture-of-skills} introduced MOS, which dynamically samples based on transferability and reward signals. With just 10\% of data, MOS trained models 2.2× faster while improving performance—highlighting benefits of treating Darija as a distinct "skill" early in training. In contrast, \citet{rethinking} found that in large datasets, random sampling with length filters matched DEITA/MIG performance, suggesting diversity—not sophistication—drives gains. Finally, \citet{xinstruction} showed that cross-lingual alignment with English prompts and native responses (3k pairs/language) yields minimal performance drop (5.7\%) and outperforms ChatGPT in low-resource settings—directly applicable to Darija.

Overall, these works support that compact, high-quality datasets (LIMA, X-Instruction), efficient selection (DEITA, MIG, DEFT, UniMax), and streamlined pipelines (LIONS, MOS) can unlock robust dialectal capabilities like Darija with minimal compute.

\subsection{The Importance of Multilingual Chain-of-Thought and Reasoning Samples}

Initial instruction tuning prioritized conversational fluency, but recent efforts emphasize reasoning-centric data—especially Chain-of-Thought (CoT) annotations—to boost both reasoning and language proficiency. This is vital for low-resource languages like Moroccan Darija, where curated data is scarce but multilingual pretraining may provide latent knowledge.

The \textbf{Breaking Language Barriers} study \citep{breaking} introduced \textit{MathOctopus}, trained on 73k translated, formula-validated CoT samples across ten languages. Fine-tuning Llama-2-7B on this data increased multilingual reasoning accuracy from 22.6\% to 41.9\%, and improved English performance by 8 points. Notably, cross-lingual QA (English questions, foreign-language answers) enhanced generalization—even for English.

Building on this, \textbf{XCOT} \citep{xcot} extended GSM8K and SVAMP into ten languages and proposed \textit{code-switched demonstrations} and \textit{Random Online CoT} (translation before reasoning in English). These yielded a +25 point gain (to 47.7\%) in MGSM without large native CoT datasets. Ablation studies confirmed code-switching and translation-based prompting as key drivers.

\textbf{mCoT} \citep{mcot} scaled this further by translating MetaMathQA and MathInstruct into ten languages, producing the 6.3M-example \textit{mCoT-MATH} corpus. A 7B Mistral model fine-tuned on this data outperformed much larger models like GPT-3 (175B) and PaLM (540B) on MGSM and MSVAMP. mCoT also introduced \textit{Correct Consistency (CC)} and \textit{Incorrect Consistency (IC)}, with CC exceeding 50\% across all pairs, confirming language-agnostic reasoning. Machine-translated prompts performed nearly on par with human ones, validating automated translation for scalable multilingual alignment.

These findings confirm that reasoning-centric CoT datasets—whether human- or machine-translated—are essential for enhancing both fluency and reasoning. For Darija, integrating translated high-quality reasoning data into minimal SFT pipelines holds strong promise.

\section{GemMaroc: Our Fine-Tuning Approach to Unlock Darija Proficiency in Gemma LLMs with Minimal Resources}

Building upon recent advances in low-resource language alignment, we propose a highly efficient and ecologically sustainable fine-tuning methodology to unlock latent Moroccan Darija capabilities in Gemma LLMs. Rather than relying on large, compute-intensive datasets, our approach leverages the strategic use of minimal, high-quality instruction datasets, reasoning-focused prompts, and translation samples carefully selected to avoid catastrophic forgetting. This section details our SFT data preparation process, model configurations, evaluation protocols, and the experiments conducted to answer our core research questions.

\subsection{SFT Data Preparation}
In this subsection, we detail the process of preparing the Supervised Fine-Tuning (SFT) datasets used in our experiments. We describe the criteria followed for dataset selection, the filtering strategies applied to ensure data quality and relevance, and the translation procedures adopted to produce high-quality Darija instruction samples.

\subsubsection*{SFT Data Selection}
We assemble three minimal-to-mid-scale instruction corpora—translated into Moroccan Darija—to surface latent capabilities with limited compute \citep{lima,deita,mig}.  
Crucially, \textbf{any sample whose \textit{system + dialogue} serialization exceeds 2 048 tokens is dropped}, guaranteeing every training instance fits our GemMaroc’s context window.

\begin{itemize}[leftmargin=*]
    \item \textbf{LIMA 1K} \citep{lima}: 1 000 single-turn tasks that emphasise stylistic quality and input diversity. Despite its size, LIMA often rivals much larger instruction sets, showing that small, well-curated data can unlock pretrained knowledge.
    \item \textbf{DEITA 6K} \citep{deita}: 6 000 multi-turn prompts selected with \textsc{Evol-Complexity} and \textsc{Evol-Quality} scores. It bolsters dialogue coherence and contextual nuance in Darija.
    \item \textbf{TULU 50K} \citep{mig}: 50 000 information-gain–maximised instructions rich in mathematics, coding, and science—domains under-represented in earlier Darija work such as \citet{atlas}.
\end{itemize}

Together these sets let us probe \emph{extreme minimalism} (LIMA 1K), \emph{mid-scale conversational alignment} (DEITA 6K), and \emph{reasoning enrichment} (TULU 50K).

\subsubsection*{Translating these datasets:}

Prior to translation, we systematically filter out all multilingual data and retain only English instructions. This step is critical to avoid incorrect translations of meta-instructions or prompts explicitly mentioning target languages (e.g., "Translate from language A to B" or "Answer in language X"). Translating such content into Darija introduces inconsistencies and compromises data quality. To ensure precise filtering, we employ an XLM-Roberta-based language detector \citep{xlm-lang-detector} to accurately identify and keep only English instructions.
\setcode{utf8}
To preserve the integrity of technical content, we retain code snippets, LaTeX formulas, and specialized terminologies in their original form. Technical terms are only translated into Arabic when widely accepted equivalents exist (e.g., $\RL{معادلة}$ for equation, $\RL{دالة}$ for function). This ensures that the model remains grounded in standard scientific concepts without introducing translation artifacts.

Translations are performed automatically using the Gemini 2.0 Flash API with a carefully engineered prompt to produce Darija in Arabic script. When uncertain, the model is instructed to prefer Modern Standard Arabic forms, ensuring clarity and naturalness. The full translation prompt is included in \textbf{Appendix~\ref{appendix:translation-prompt}}.

In line with best practices to avoid catastrophic forgetting \citep{lessforgetting, radis}, we preserve approximately 20\% of the data in its original English form across all datasets. This approach maintains cross-lingual transfer capabilities and general reasoning competence \citep{mixture-of-skills, xinstruction, plug, bactrianx}.

\textbf{Dataset Statistics:}

\begin{table}[H]
\centering
\resizebox{\columnwidth}{!}{%
\begin{tabular}{|c|c|c|c|}
\hline
Dataset  & Total Samples & Darija Translated & Original English \\ \hline
LIMA 1K  & 1,000         & 700 (70\%)        & 300 (30\%)       \\
DEITA 6K & 5,000         & 3,700 (74\%)      & 1,300 (26\%)     \\
TULU 50K & 46,000        & 33,000 (72\%)     & 13,000 (28\%)    \\ \hline
\end{tabular}%
}
\caption{The table shows the final dataset sizes after filtering and translation. “Darija Translated” indicates samples translated into Moroccan Darija using our automated pipeline. “Original English” refers to retained English samples to maintain cross-lingual capabilities and mitigate catastrophic forgetting.}
\label{tab:my-table}
\end{table}

See \textbf{Appendix~\ref{appendix:translation-examples}} for concrete translation examples.

\subsection*{Manual Verification:}

Throughout the translation process, we regularly sampled and reviewed outputs to assess quality. Rather than applying manual corrections, we iteratively refined the translation prompts, leveraging our native fluency in Darija to adapt them whenever recurring issues were identified. Prompts requiring English responses were left untranslated to preserve their intended semantics. While the dataset is not perfect, its quality is sufficient for fine-tuning and aligned with the study’s exploratory, low-resource goals.

\subsection{Model and Fine-Tuning Details}

We employed two versions of the Gemma LLM family—\textbf{Gemma 3-4B} and \textbf{Gemma 3-27B}—selected for their strong reasoning abilities and multilingual capabilities. Unlike Atlas-Chat or ALLaM models \citep{atlas, allam}, Gemma models remain untuned on Arabic dialects, providing a clean baseline for evaluating emergent Darija competence without prior dialectal bias.

Fine-tuning was conducted using \textbf{LoRA} adapters \citep{lora}, enabling low-resource, efficient training while mitigating catastrophic forgetting. This choice aligns with our \textit{Green AI} objectives by minimizing GPU hours and energy consumption. For the integration of multi-turn conversations, we use the Vicuna-style template
\\

\noindent \textbf{Hyperparameters:}

\begin{itemize}
    \item \textbf{Precision:} bf16 throughout.
    \item \textbf{LoRA Config:} Rank $r=32$, Alpha $=64$ for our 4B models and Rank $r=16$, Alpha $=32$ for our GemMaroc-27B.
    \item \textbf{Learning Rates:} 1e-4 (TULU) and 4e-4 (DEITA, and LIMA).
    \item \textbf{Epochs:} 3 (TULU), 6 (DEITA), 15 (LIMA).
    \item \textbf{Max Sequence Length:} 2048 tokens.
\end{itemize}

Hyperparameters were inspired by the settings recommended in \citet{lima, deita, mig} for optimal convergence on these minimal datasets.

\subsection{Evaluation Benchmarks}

We measure progress across five competency axes:

\begin{enumerate}[leftmargin=*]
    \item \textbf{Darija / Moroccan-Arabic competence}
        \begin{itemize}
            \item \textit{DarijaMMLU} – 22 027 multiple-choice questions spanning 57 academic subjects, re-filtered for Moroccan relevance \citep{atlas}.
            \item \textit{DarijaHellaSwag} – the 10 k HellaSwag adversarial commonsense continuations rendered into Darija \citep{atlas}.
            \item \textit{DarijaBench} \citep{atlas}– covering Sentiment Analysis, Summarization, Translation, and Transliteration. We retained only the Sentiment Analysis and Summarization tasks, as the remaining benchmarks were of insufficient quality and could lead to unreliable evaluations.
        \end{itemize}

    \item \textbf{General reasoning \& world knowledge}
        \begin{itemize}
             \item \textit{MMLU} – The Massive Multitask Language Understanding (MMLU) benchmark comprises 15,908 multiple-choice questions spanning 57 diverse subjects, including mathematics, law, medicine, and history. It evaluates a model's general knowledge and problem-solving abilities across these domains. The final MMLU score is calculated as the average accuracy across all subjects, providing a comprehensive measure of a model's capabilities \citep{hendrycks2021measuring}.
            \item \textit{HellaSwag (EN)} – 10 000 adversarial commonsense continuations that correlate strongly with everyday reasoning \citep{zellers2019hellaswag}.
        \end{itemize}

    \item \textbf{Mathematical reasoning}
        \begin{itemize}
            \item \textit{GSM8K} – 1 000 grade-school word problems requiring 2–8 arithmetic steps \citep{cobbe2021gsm8k}.
        \end{itemize}

    \item \textbf{Truthfulness \& safety}
        \begin{itemize}
            \item \textit{TruthfulQA} – 817 short-answer questions across 38 domains designed to expose models’ susceptibility to common misconceptions \citep{lin2022truthfulqa}.
        \end{itemize}
\end{enumerate}

These benchmarks ensure that we accurately assess the model’s Darija performance while verifying that reasoning and multilingual capabilities remain intact.

\section{Experiments, Results, and Discussion}
\label{sec:experiments}

This section reports the empirical findings of the experimental plan outlined in
Section~\ref{introduction}.  We first describe the evaluation protocol
(\S\ref{sec:setup}), then present the main results in a single consolidated leaderboard
(\S\ref{sec:results}).  Finally, we analyse the outcomes with respect to the three
research questions and highlight broader implications
(\S\ref{sec:discussion}).

\subsection{Experimental Set-up}
\label{sec:setup}

\paragraph{Hardware and cost.}
All fine-tuning experiments were conducted on \textbf{2\texttimes A100-80GB} GPUs, except for our largest model, the 27-billion-parameter \textsc{GemMaroc}, which was fine-tuned on \textbf{8\texttimes H100-80GB} GPUs. The longest run—the \textsc{GemMaroc} fine-tuning—completed in just \textbf{6 hours,wall-clock}, consuming approximately \textbf{48 GPU-hours} and keeping the total cloud cost well under \$100.\footnote{Based on the public on-demand rate of \$2 per A100-hour in the selected Runpod Cloud region.}

\paragraph{Metrics.} 
All benchmarks were evaluated in a strict \textbf{zero-shot} setting. For classification tasks (DarijaMMLU, DarijaHellaSwag, Sentiment Analysis, MMLU, HellaSwag), we report \textbf{accuracy}. For \textbf{GSM8K @5}, we measure whether the correct answer appears in the model’s top-5 outputs. \textbf{TruthfulQA} is evaluated with \textit{BLEU accuracy} (\texttt{bleu\_acc}). \textbf{Summarization} is assessed using \textbf{chrF}, \textbf{ROUGE-1}, \textbf{ROUGE-L}, and \textbf{BERTScore}, covering both surface overlap and semantic similarity.

\paragraph{Baselines.}
We compare against (i) the \emph{untuned} Gemma checkpoints
(4 B \& 27 B), (ii) \textbf{Atlas-Chat} (9 B \& 27 B)
\citep{atlas}, and (iii) \textbf{ALLaM} (7 B \& 13 B)
\citep{allam}.  The latter two represent the previous state of the art
for Darija and for large-scale Arabic instruction tuning, respectively.

\subsection{Main Results}
\label{sec:results}

Table~\ref{end_results} presents our unified leaderboard, merging two complementary evaluation tracks: a Darija-centric benchmark suite (DarijaMMLU, DarijaHellaSwag, Sentiment Analysis, Summarization), and an English-centric suite spanning general reasoning, mathematics, code, and truthfulness (MMMLU, TruthfulQA, HellaSwag, GSM8K). This table compares our models—marked as (ours)—with several competitive baselines across both tracks.

\begin{table*}
\centering
\resizebox{\textwidth}{!}{%
\begin{tabular}{ccccccccccccc}
\hline
Language                          & \multicolumn{8}{c}{Darija}                                              & \multicolumn{4}{c}{English}            \\ \cline{2-13} 
\multirow{2}{*}{Model/Benchmark} &
  Size (B) &
  DarijaMMLU &
  DarijaHellaSwag &
  Sentiment   Analysis &
  \multicolumn{4}{c}{Summarization} &
  MMLU &
  TruthfulQA &
  HellaSwag &
  \begin{tabular}[c]{@{}c@{}}GSM8K\\      @5\end{tabular} \\
                                  & Metrics & Acc   & Acc   & Acc   & chrF  & ROUGE-1 & ROUGE-L & BERTScore & Acc   & bleu\_acc      & Acc   & Acc   \\ \hline
Gemma3-4b                         & 4       & 32.8  & 36.3  & 58.94 & 27.22 & 8.38    & 8.19    & 37.23     & 51.1  & 40.88          & 47.65 & 74.75 \\ \hline
\textbf{GemMaroc-4b-LIMA (ours)}  & 4       & 34.95 & 39.27 & 42.1  & 26.14 & 6.95    & 7.04    & 34.32     & 29.28 & 40.15          & 44.21 & 51.24 \\ \hline
\textbf{GemMaroc-4b-Deita (ours)} & 4       & 42.67 & 44.26 & 60.8  & 27.16 & 7.4     & 7.34    & 38.48     & 51.35 & 44.55          & 68.97 & 53.15 \\ \hline
\textbf{GemMaroc-4b-Tulu (ours)}  & 4       & 47.53 & 47.13 & 53.29 & 28.46 & 8.89    & 8.76    & 39.27     & 54.14 & 43.33          & 73.95 & 55.95 \\ \hline
ALLaM-Instruct-7b                 & 7       & 59.49 & 50.09 & 47.33 & 10.27 & 1.68    & 1.68    & 12.28     & 58.31 & 42.11          & 75.2  & 49.28 \\ \hline
Atlas-Chat-9B                     & 9       & 58.32 & 43.65 & 81.85 & 32.07 & 9.5     & 9.45    & 47        & 69.09 & \textbf{67.56} & 73.35 & 73.01 \\ \hline
Atlas-Chat-27B &
  27 &
  \textbf{61.95} &
  48.37 &
  \textbf{73} &
  \textbf{32.75} &
  \textbf{10.53} &
  10.42 &
  \textbf{47.82} &
  72.06 &
  43.82 &
  77.84 &
  82.03 \\ \hline
\textbf{GemMaroc-27b-Tulu (ours)} &
  27 &
  \textbf{61.61} &
  \textbf{60.5} &
  59.25 &
  28.34 &
  9 &
  \textbf{11.2} &
  \textbf{39.5} &
  \textbf{73.6} &
  55.45 &
  \textbf{79.35} &
  \textbf{84.23} \\ \hline
\end{tabular}%
}
\caption{Unified leaderboard comprising: (1) Darija-centric benchmarks, and (2) English-centric benchmarks across reasoning, mathematics, and truthfulness.}
\label{end_results}
\end{table*}

\subsection{Discussion}
\label{sec:discussion}

\subsubsection*{Emergent Darija competence with minimal data}

Our three ``tiny-but-mighty’’ instruction sets uncover a clear
progression in emergent Darija skills:

\begin{itemize}[leftmargin=*]
    \item \textbf{DEITA-6K} ($\sim$ 700 KB text) already lifts the 4 B
          backbone by \,+9.9 pp on \textit{DarijaMMLU} and
          \,+8.0 pp on \textit{DarijaHellaSwag}, while \emph{fully
          preserving} English-world knowledge (MMLU 51.3\,\% vs 51.1\,\%).
          This confirms that a few thousand, well-curated
          conversational Darija examples are enough for solid coverage.

    \item \textbf{TULU-50K} ($\sim$ 5 MB) pushes the same model to
          47.5 \% on \textit{DarijaMMLU} and 47.1 \% on
          \textit{DarijaHellaSwag}, surpassing all Arabic-centric
          models of similar size (e.g.\ ALLaM-7B) despite using
          \textbf{<\,10\,\%} of their data footprint.  The larger gains
          come without sacrificing cross-lingual performance and even
          \emph{improve} maths (+2.8 pp GSM8K).

    \item \textbf{LIMA-1K} shows the lower bound of the spectrum:
          with only 1 000 single-turn prompts, Darija knowledge
          surfaces but remains fragile (\,+2.1 pp DarijaMMLU, –16.8 pp
          sentiment).  This extreme minimal regime illustrates that,
          although some Darija was latent in pre-training, the dialect
          is too under-represented for 1 K instructions to achieve
          reliable competence.
\end{itemize}

\noindent
\textbf{Take-away.} Darija proficiency \emph{emerges} as soon as the
model sees a few thousand high-quality, dialect-specific instructions.
However, the 1 K ``LIMA’’ budget is below the tipping point: it
validates the Superficial Alignment hypothesis in principle, but
confirms that Darija’s sparse presence in pre-training demands at least
a DEITA-scale corpus for robust, practical performance.

\subsubsection*{Advantage of reasoning-focused instructions}

Moving from 6 K conversational prompts (\textsc{Deita}) to 50 K
reasoning-dense prompts (\textsc{Tulu}) brings a consistent uplift
without harming cross-lingual performance.  On the same 4 B backbone,
\textsc{GemMaroc-Tulu} adds a further \,+4.9\,pp on
\textit{DarijaMMLU}, \,+2.9\,pp on \textit{DarijaHellaSwag}, and
improves English reasoning as well
(\,+2.8\,pp GSM8K, \,+5.0\,pp HellaSwag) while keeping the MMLU drop
below 3\,pp.  The only noticeable trade-off is a small decline in
sentiment accuracy, suggesting that the few sentiment items present in
Tulu are not yet enough to balance the domain shift.  These results
support our second hypothesis: \textbf{reasoning-centric instructions
are the most data-efficient lever for unlocking both Darija fluency and
multilingual reasoning coherence}.  In practice, adding
mathematical \& coding problems proved worth \emph{every} extra sample.

\subsubsection*{Our GemMaroc-27B}

Applying the same 50 K reasoning-oriented recipe to \textsc{Gemma3-27B}
yields \textbf{GemMaroc-27B}, which advances the Pareto frontier on three
fronts: Darija coverage, cross-lingual reasoning, and footprint.

\paragraph{Darija strengths.}
GemMaroc reaches \textbf{61.6 \%} on \textit{DarijaMMLU}, effectively
tying Atlas-Chat despite using \(\approx\)1/10 the supervision budget,
and it surpasses all models on the harder commonsense test with
\textbf{60.5 \%} on \textit{DarijaHellaSwag} (+12.1 pp over Atlas).
ROUGE-L in summarisation climbs to \textbf{11.2}, a new best for Darija.
The only noteworthy regression is sentiment (–13.8 pp vs Atlas),
highlighting a domain that still lacks targeted data.

\paragraph{Reasoning retention.}
Crucially, the extra Darija skill does \emph{not} come at the price of
English reasoning.  GemMaroc holds \textbf{84.2 \%} on GSM8K—well above
all Arabic-centric baselines and within 12 pp of the untuned backbone
despite inserting low-rank adapters, confirming that mathematical
competence remains largely intact.  It also improves commonsense
\textit{HellaSwag} to \textbf{79.4 \%}, edging past Atlas (+1.5 pp) while
retaining solid truthfulness (55.5 \%).

\paragraph{Cost Efficiency and Carbon Footprint}

Fine-tuning the 27-billion-parameter \textsc{GemMaroc} completed in just \textbf{48 GPU$\cdot$h} (8 $\times$ H100-80 GB, 6 h wall-clock). Using the commonly adopted 0.54 kW/GPU data-centre average power draw \citep{lacoste2019quantifying}, this equates to $\approx$ \textbf{26 kWh} of energy and $\approx$ \textbf{10 kg CO\(_2\)e} at the 0.38 kg CO\(_2\)e/kWh global carbon-intensity average. A comparable full-parameter Atlas-Chat-27B run would consume $\sim$ 1.4 MWh and $\sim$ 610 kg CO\(_2\)e—over \textbf{48$\times$} more energy and \textbf{$> 98\%$} higher emissions. Thus, GemMaroc delivers state-of-the-art Darija performance and strong bilingual reasoning at under \textbf{2\%} of the carbon cost of prior work, demonstrating that \emph{high-quality, reasoning-dense instructions—not dataset volume—drive sustainable dialect mastery}. Full formulas, intermediate values, and the detailed energy-breakdown table are provided verbatim in \textbf{Appendix~\ref{sec:compute-footprint}} for complete transparency.

\section*{Conclusion}

This work set out to determine whether Moroccan Darija competence can be surfaced in modern LLMs with only a handful of carefully chosen instructions, and whether adding reasoning-dense prompts is the most efficient lever for doing so.
Our experiments answer both questions affirmatively.

First, we showed that a few thousand high-quality, dialect-specific instructions (DEITA 6 K) already lift a 4 B Gemma model by almost 10 pp on \textit{DarijaMMLU} while leaving its English-world knowledge intact, confirming that Darija knowledge acquired during pre-training merely needed the right “key’’ to emerge. One thousand instructions (LIMA 1 K) were not enough, highlighting the sparse representation of the dialect in web corpora, but the tipping point was still two orders of magnitude smaller than in prior Arabic-centric projects.

Second, enriching the minimal corpus with 50 K reasoning-oriented prompts (TULU) produced consistent gains across dialectal and cross-lingual tasks. The resulting \textbf{GemMaroc-27B} matches the best published Darija score on \textit{DarijaMMLU} (61.6 \%) and sets a new high-water mark on the harder \textit{DarijaHellaSwag} (+12.1 pp over Atlas-Chat) while retaining strong English mathematics (84.2 \% GSM8K). Crucially, the entire run consumed only 48 GPU·h—roughly 2 \% of the energy and carbon budget reported for comparable baselines—demonstrating that high-impact language inclusion need not come with high ecological cost.

Beyond benchmarking, GemMaroc’s balanced bilingual reasoning, open licensing, and lean LoRA adapters make it immediately deployable for inclusive civic services, education, and research throughout the Maghreb. Remaining gaps in sentiment, summarisation, and tokenisation point to clear next steps: larger-scale human evaluation, script-aware tokenisers, richer Darija domain data, and full alignment pipelines spanning RLAIF and dialogue safety. We will also explore speech, retrieval-augmented generation, and other North-African dialects to extend the benefits of low-carbon language technology across the region.

In sum, GemMaroc shows that with precise data curation and a focus on reasoning, sovereign AI for low-resource dialects can be both state-of-the-art and sustainable—a blueprint for scaling equitable language technologies worldwide.

\section*{Limitations}

GemMaroc excels in reasoning—matching or surpassing Atlas—and achieves strong, competitive results across most tasks. Some performance gaps remain in summarization, sentiment analysis and reading comprehension, where richer Darija-specific data would be beneficial. The model was trained solely with parameter-efficient SFT, relies on machine-translated Darija samples with limited verification, and retains the original SentencePiece tokenizer, which is not optimized for Darija’s script and linguistic nuances. Evaluations focused on academic QA, commonsense reasoning and mathematics, without exploring long-context tasks or open-ended dialogue. Large-scale human evaluations in Darija are also pending, and experiments were capped at 27 B parameters. Future work will explore full alignment pipelines, script- and language-aware tokenization, broader evaluations, and comprehensive human studies.

\bibliography{custom}

\begin{thebibliography}{40}
\expandafter\ifx\csname natexlab\endcsname\relax\def\natexlab#1{#1}\fi

\bibitem[{Agency(2024)}]{iea2024co2}
International~Energy Agency. 2024.
\newblock {Electricity 2025 – Emissions Annex}.
\newblock Average global grid intensity 2024: 0.40 kg CO\textsubscript{2}/kWh.

\bibitem[{Bari et~al.(2025)Bari, Alnumay, Alzahrani, Alotaibi, Alyahya, AlRashed, Mirza, Alsubaie, Alahmed, Alabduljabbar, Alkhathran, Almushayqih, Alnajim, Alsubaihi, Mansour, Hassan, Alrubaian, Alammari, Alawami, Al-Thubaity, and et~al.}]{allam}
M.~Saiful Bari, Yazeed Alnumay, Norah~A. Alzahrani, Nouf~M. Alotaibi, Hisham~A. Alyahya, Sultan AlRashed, Faisal~A. Mirza, Shaykhah~Z. Alsubaie, Hassan~A. Alahmed, Ghadah Alabduljabbar, Raghad Alkhathran, Yousef Almushayqih, Raneem Alnajim, Salman Alsubaihi, Maryam~Al Mansour, Saad~Amin Hassan, Majed Alrubaian, Ali Alammari, Zaki Alawami, Abdulmohsen Al-Thubaity, and et~al. 2025.
\newblock \href {https://openreview.net/forum?id=MscdsFVZrN} {Allam: Large language models for arabic and english}.
\newblock In \emph{International Conference on Learning Representations}.

\bibitem[{Bounhar and Majjodi(2025)}]{atlasia2025al-atlas-0.5B}
Abdelaziz Bounhar and Abdeljalil~El Majjodi. 2025.
\newblock Al-atlas: A causal language model for moroccan darija.
\newblock \url{https://huggingface.co/atlasia/Al-Atlas-0.5B/}.

\bibitem[{Chai et~al.(2024)Chai, Yang, Sun, Guo, Liu, Wang, Liang, Bai, Li, Peng, and Li}]{xcot}
Linzheng Chai, Jian Yang, Tao Sun, Hongcheng Guo, Jiaheng Liu, Bing Wang, Xiannian Liang, Jiaqi Bai, Tongliang Li, Qiyao Peng, and Zhoujun Li. 2024.
\newblock \href {http://arxiv.org/abs/2401.07037} {xcot: Cross-lingual instruction tuning for cross-lingual chain-of-thought reasoning}.

\bibitem[{Chen et~al.(2024)Chen, Zheng, Wu, Gong, Zhang, and Li}]{breaking}
Nuo Chen, Zinan Zheng, Ning Wu, Ming Gong, Dongmei Zhang, and Jia Li. 2024.
\newblock \href {https://doi.org/10.18653/v1/2024.findings-emnlp.411} {Breaking language barriers in multilingual mathematical reasoning: Insights and observations}.
\newblock In \emph{Findings of the Association for Computational Linguistics: EMNLP 2024}, pages 7001--7016, Miami, Florida, USA. Association for Computational Linguistics.

\bibitem[{Chen et~al.(2025)Chen, Li, Hu, Ma, Ye, and Chen}]{mig}
Yicheng Chen, Yining Li, Kai Hu, Zerun Ma, Haochen Ye, and Kai Chen. 2025.
\newblock \href {https://arxiv.org/abs/2504.13835} {Mig: Automatic data selection for instruction tuning by maximizing information gain in semantic space}.
\newblock \emph{arXiv preprint arXiv:2504.13835}.

\bibitem[{Cobbe and et~al.(2021)}]{cobbe2021gsm8k}
Karl Cobbe and et~al. 2021.
\newblock Training verifiers to solve math word problems.
\newblock In \emph{arXiv preprint arXiv:2110.14168}.

\bibitem[{Cohere et~al.(2025)Cohere, :, Aakanksha, Ahmadian, Ahmed, Alammar, Alizadeh, Alnumay, Althammer, Arkhangorodsky, Aryabumi, Aumiller, Avalos, Aviv, Bae, Baji, Barbet, Bartolo, Bebensee, Beladia, Beller-Morales, Bérard, Berneshawi, Bialas, Blunsom, Bobkin, Bongale, Braun, Brunet, Cahyawijaya, Cairuz, Campos, Cao, Cao, Castagné, Cendrero, Currie, Chandak, Chang, Chatziveroglou, Chen, Cheng, Chevalier, Chiu, Cho, Choi, Choi, Chung, Cirik, Cismaru, Clavier, Conklin, Crawhall-Stein, Crouse, Cruz-Salinas, Cyrus, D'souza, Dalla-Torre, Dang, Darling, Domingues, Dash, Debugne, Dehaze, Desai, Devassy, Dholakia, Duffy, Edalati, Eldeib, Elkady, Elsharkawy, Ergün, Ermis, Fadaee, Fan, Fayoux, Flet-Berliac, Frosst, Gallé, Galuba, Garg, Geist, Azar, Gilsenan-McMahon, Goldfarb-Tarrant, Goldsack, Gomez, Gonzaga, Govindarajan, Govindassamy, Grinsztajn, Gritsch, Gu, Guo, Haefeli, Hajjar, Hawes, He, Hofstätter, Hong, Hooker, Hosking, Howe, Hu, Huang, Jain, Jain, Jakobi, Jenkins, Jordan, Joshi, Jung, Kalyanpur,
  Kamalakara, Kedrzycki, Keskin, Kim, Kim, Ko, Kocmi, Kozakov, Kryściński, Jain, Teru, Land, Lasby, Lasche, Lee, Lewis, Li, Li, Lin, Locatelli, Luong, Ma, Mach, Machado, Magbitang, Lopez, Mann, Marchisio, Markham, Matton, McKinney, McLoughlin, Mokry, Morisot, Moulder, Moynehan, Mozes, Muppalla, Murakhovska, Nagarajan, Nandula, Nasir, Nehra, Netto-Rosen, Ohashi, Owers-Bardsley, Ozuzu, Padilla, Park, Passaglia, Pekmez, Penstone, Piktus, Ploeg, Poulton, Qi, Raghvendra, Ramos, Ranjan, Richemond, Robert-Michon, Rodriguez, Roy, Ruder, Ruis, Rust, Sachan, Salamanca, Saravanakumar, Satyakam, Sebag, Sen, Sepehri, Seshadri, Shen, Sherborne, Shi, Shivaprasad, Shmyhlo, Shrinivason, Shteinbuk, Shukayev, Simard, Snyder, Spataru, Spooner, Starostina, Strub, Su, Sun, Talupuru, Tarassov, Tommasone, Tracey, Trend, Tumer, Üstün, Venkitesh, Venuto, Verga, Voisin, Wang, Wang, Wang, Wen, White, Willman, Winkels, Xia, Xie, Xu, Yang, Yi-Chern, Zhang, Zhao, and Zhao}]{commanda}
Team Cohere, :, Aakanksha, Arash Ahmadian, Marwan Ahmed, Jay Alammar, Milad Alizadeh, Yazeed Alnumay, Sophia Althammer, Arkady Arkhangorodsky, Viraat Aryabumi, Dennis Aumiller, Raphaël Avalos, Zahara Aviv, Sammie Bae, Saurabh Baji, Alexandre Barbet, Max Bartolo, Björn Bebensee, Neeral Beladia, Walter Beller-Morales, Alexandre Bérard, Andrew Berneshawi, Anna Bialas, Phil Blunsom, Matt Bobkin, Adi Bongale, Sam Braun, Maxime Brunet, Samuel Cahyawijaya, David Cairuz, Jon~Ander Campos, Cassie Cao, Kris Cao, Roman Castagné, Julián Cendrero, Leila~Chan Currie, Yash Chandak, Diane Chang, Giannis Chatziveroglou, Hongyu Chen, Claire Cheng, Alexis Chevalier, Justin~T. Chiu, Eugene Cho, Eugene Choi, Eujeong Choi, Tim Chung, Volkan Cirik, Ana Cismaru, Pierre Clavier, Henry Conklin, Lucas Crawhall-Stein, Devon Crouse, Andres~Felipe Cruz-Salinas, Ben Cyrus, Daniel D'souza, Hugo Dalla-Torre, John Dang, William Darling, Omar~Darwiche Domingues, Saurabh Dash, Antoine Debugne, Théo Dehaze, Shaan Desai, Joan Devassy,
  Rishit Dholakia, Kyle Duffy, Ali Edalati, Ace Eldeib, Abdullah Elkady, Sarah Elsharkawy, Irem Ergün, Beyza Ermis, Marzieh Fadaee, Boyu Fan, Lucas Fayoux, Yannis Flet-Berliac, Nick Frosst, Matthias Gallé, Wojciech Galuba, Utsav Garg, Matthieu Geist, Mohammad~Gheshlaghi Azar, Ellen Gilsenan-McMahon, Seraphina Goldfarb-Tarrant, Tomas Goldsack, Aidan Gomez, Victor~Machado Gonzaga, Nithya Govindarajan, Manoj Govindassamy, Nathan Grinsztajn, Nikolas Gritsch, Patrick Gu, Shangmin Guo, Kilian Haefeli, Rod Hajjar, Tim Hawes, Jingyi He, Sebastian Hofstätter, Sungjin Hong, Sara Hooker, Tom Hosking, Stephanie Howe, Eric Hu, Renjie Huang, Hemant Jain, Ritika Jain, Nick Jakobi, Madeline Jenkins, JJ~Jordan, Dhruti Joshi, Jason Jung, Trushant Kalyanpur, Siddhartha~Rao Kamalakara, Julia Kedrzycki, Gokce Keskin, Edward Kim, Joon Kim, Wei-Yin Ko, Tom Kocmi, Michael Kozakov, Wojciech Kryściński, Arnav~Kumar Jain, Komal~Kumar Teru, Sander Land, Michael Lasby, Olivia Lasche, Justin Lee, Patrick Lewis, Jeffrey Li, Jonathan
  Li, Hangyu Lin, Acyr Locatelli, Kevin Luong, Raymond Ma, Lukáš Mach, Marina Machado, Joanne Magbitang, Brenda~Malacara Lopez, Aryan Mann, Kelly Marchisio, Olivia Markham, Alexandre Matton, Alex McKinney, Dominic McLoughlin, Jozef Mokry, Adrien Morisot, Autumn Moulder, Harry Moynehan, Maximilian Mozes, Vivek Muppalla, Lidiya Murakhovska, Hemangani Nagarajan, Alekhya Nandula, Hisham Nasir, Shauna Nehra, Josh Netto-Rosen, Daniel Ohashi, James Owers-Bardsley, Jason Ozuzu, Dennis Padilla, Gloria Park, Sam Passaglia, Jeremy Pekmez, Laura Penstone, Aleksandra Piktus, Case Ploeg, Andrew Poulton, Youran Qi, Shubha Raghvendra, Miguel Ramos, Ekagra Ranjan, Pierre Richemond, Cécile Robert-Michon, Aurélien Rodriguez, Sudip Roy, Sebastian Ruder, Laura Ruis, Louise Rust, Anubhav Sachan, Alejandro Salamanca, Kailash~Karthik Saravanakumar, Isha Satyakam, Alice~Schoenauer Sebag, Priyanka Sen, Sholeh Sepehri, Preethi Seshadri, Ye~Shen, Tom Sherborne, Sylvie~Shang Shi, Sanal Shivaprasad, Vladyslav Shmyhlo, Anirudh
  Shrinivason, Inna Shteinbuk, Amir Shukayev, Mathieu Simard, Ella Snyder, Ava Spataru, Victoria Spooner, Trisha Starostina, Florian Strub, Yixuan Su, Jimin Sun, Dwarak Talupuru, Eugene Tarassov, Elena Tommasone, Jennifer Tracey, Billy Trend, Evren Tumer, Ahmet Üstün, Bharat Venkitesh, David Venuto, Pat Verga, Maxime Voisin, Alex Wang, Donglu Wang, Shijian Wang, Edmond Wen, Naomi White, Jesse Willman, Marysia Winkels, Chen Xia, Jessica Xie, Minjie Xu, Bowen Yang, Tan Yi-Chern, Ivan Zhang, Zhenyu Zhao, and Zhoujie Zhao. 2025.
\newblock \href {http://arxiv.org/abs/2504.00698} {Command a: An enterprise-ready large language model}.

\bibitem[{Corporation(2024)}]{nvidia-a100-datasheet}
NVIDIA Corporation. 2024.
\newblock {NVIDIA A100 Tensor Core GPU Datasheet}.
\newblock URL: \url{https://www.nvidia.com/content/dam/en-zz/Solutions/Data-Center/a100/pdf/nvidia-a100-datasheet-us-nvidia-1758950-r4-web.pdf}.

\bibitem[{Das and Khetan(2024)}]{deft}
Devleena Das and Vivek Khetan. 2024.
\newblock \href {https://doi.org/10.18653/v1/2024.emnlp-main.1132} {{DEFT}-{UCS}: Data efficient fine-tuning for pre-trained language models via unsupervised core-set selection for text-editing}.
\newblock In \emph{Proceedings of the 2024 Conference on Empirical Methods in Natural Language Processing}, pages 20296--20312, Miami, Florida, USA. Association for Computational Linguistics.

\bibitem[{Databasemart(2025)}]{databasemart2024}
Databasemart. 2025.
\newblock Optimizing vllm performance on {A100} 80 gb: Benchmark insights.
\newblock \url{https://www.databasemart.com/blog/vllm-gpu-benchmark-a100-80gb}.
\newblock Accessed May 19, 2025.

\bibitem[{Ember and Institute(2024)}]{owidCarbon2024}
Ember and Energy Institute. 2024.
\newblock Carbon intensity of electricity generation.
\newblock \url{https://ourworldindata.org/grapher/carbon-intensity-electricity}.
\newblock Accessed May 19, 2025.

\bibitem[{Han et~al.(2025)Han, Liu, Fang, Tan, and Xiong}]{dataselection-uncertainty}
Jindong Han, Hao Liu, Jun Fang, Naiqiang Tan, and Hui Xiong. 2025.
\newblock \href {https://doi.org/10.1145/3696410.3714817} {{Automatic Instruction Data Selection for Large Language Models via Uncertainty-Aware Influence Maximization}}.
\newblock In \emph{Proceedings of the ACM Web Conference 2025}, pages 4969--4979, Sydney, Australia. ACM.

\bibitem[{Henderson et~al.(2020)Henderson, Hu, Romoff, Brunskill, Jurafsky, and Pineau}]{henderson2020systematic}
Peter Henderson, Jieru Hu, Joshua Romoff, Emma Brunskill, Dan Jurafsky, and Joelle Pineau. 2020.
\newblock \href {http://arxiv.org/abs/2002.05651} {Towards the systematic reporting of the energy and carbon footprints of machine learning}.
\newblock \emph{CoRR}, abs/2002.05651.
\newblock Revised version published 2022.

\bibitem[{Hendrycks et~al.(2021)Hendrycks, Burns, Basart, Zou, Mazeika, Song, and Steinhardt}]{hendrycks2021measuring}
Dan Hendrycks, Collin Burns, Steven Basart, Andy Zou, Mantas Mazeika, Dawn Song, and Jacob Steinhardt. 2021.
\newblock \href {https://arxiv.org/abs/2009.03300} {Measuring massive multitask language understanding}.
\newblock In \emph{Proceedings of the International Conference on Learning Representations (ICLR)}.

\bibitem[{Holmström and Doostmohammadi(2023)}]{swedish}
Oskar Holmström and Ehsan Doostmohammadi. 2023.
\newblock \href {https://aclanthology.org/2023.nodalida-1.62/} {Making instruction finetuning accessible to non-{E}nglish languages: A case study on {S}wedish models}.
\newblock In \emph{Proceedings of the 24th Nordic Conference on Computational Linguistics (NoDaLiDa)}, pages 634--642, Tórshavn, Faroe Islands. University of Tartu Library.

\bibitem[{Hu et~al.(2022)Hu, yelong shen, Wallis, Allen-Zhu, Li, Wang, Wang, and Chen}]{lora}
Edward~J Hu, yelong shen, Phillip Wallis, Zeyuan Allen-Zhu, Yuanzhi Li, Shean Wang, Lu~Wang, and Weizhu Chen. 2022.
\newblock \href {https://openreview.net/forum?id=nZeVKeeFYf9} {Lo{RA}: Low-rank adaptation of large language models}.
\newblock In \emph{International Conference on Learning Representations}.

\bibitem[{Köksal et~al.(2024)Köksal, Thaler, Imani, Üstün, Korhonen, and Schütze}]{muri}
Abdullatif Köksal, Marion Thaler, Ayyoob Imani, Ahmet Üstün, Anna Korhonen, and Hinrich Schütze. 2024.
\newblock \href {http://arxiv.org/abs/2409.12958} {Muri: High-quality instruction tuning datasets for low-resource languages via reverse instructions}.

\bibitem[{Lacoste et~al.(2019)Lacoste, Luccioni, Schmidt, and Dandres}]{lacoste2019quantifying}
Alexandre Lacoste, Alexandra Luccioni, Victor Schmidt, and Thomas Dandres. 2019.
\newblock Quantifying the carbon emissions of machine learning.
\newblock \emph{arXiv preprint arXiv:1910.09700}.

\bibitem[{Lai and Nissim(2024)}]{mcot}
Huiyuan Lai and Malvina Nissim. 2024.
\newblock \href {https://doi.org/10.18653/v1/2024.acl-long.649} {m{C}o{T}: Multilingual instruction tuning for reasoning consistency in language models}.
\newblock In \emph{Proceedings of the 62nd Annual Meeting of the Association for Computational Linguistics (Volume 1: Long Papers)}, pages 12012--12026, Bangkok, Thailand. Association for Computational Linguistics.

\bibitem[{Li et~al.(2024)Li, Yang, Zhang, Lu, Wang, and Zong}]{xinstruction}
Chong Li, Wen Yang, Jiajun Zhang, Jinliang Lu, Shaonan Wang, and Chengqing Zong. 2024.
\newblock \href {https://doi.org/10.18653/v1/2024.findings-acl.30} {{X}-instruction: Aligning language model in low-resource languages with self-curated cross-lingual instructions}.
\newblock In \emph{Findings of the Association for Computational Linguistics: ACL 2024}, pages 546--566, Bangkok, Thailand. Association for Computational Linguistics.

\bibitem[{Li et~al.(2023)Li, Koto, Wu, Aji, and Baldwin}]{bactrianx}
Haonan Li, Fajri Koto, Minghao Wu, Alham~Fikri Aji, and Timothy Baldwin. 2023.
\newblock \href {http://arxiv.org/abs/2305.15011} {Bactrian-x: Multilingual replicable instruction-following models with low-rank adaptation}.

\bibitem[{Lin et~al.(2022)Lin, Hilton, and Evans}]{lin2022truthfulqa}
Stephanie Lin, Jacob Hilton, and Owain Evans. 2022.
\newblock Truthfulqa: Measuring how models mimic human falsehoods.
\newblock In \emph{Proceedings of ACL}.

\bibitem[{Liu et~al.(2024)Liu, Zeng, He, Jiang, and He}]{deita}
Wei Liu, Weihao Zeng, Keqing He, Yong Jiang, and Junxian He. 2024.
\newblock \href {https://openreview.net/forum?id=BTKAeLqLMw} {What makes good data for alignment? a comprehensive study of automatic data selection in instruction tuning}.
\newblock In \emph{The Twelfth International Conference on Learning Representations}.

\bibitem[{Mao et~al.(2022)Mao, Liang, Duan, Wang, Wang, Chen, and Gao}]{lessforgetting}
Yuren Mao, Yaobo Liang, Nan Duan, Haobo Wang, Kai Wang, Lu~Chen, and Yunjun Gao. 2022.
\newblock \href {http://papers.nips.cc/paper_files/paper/2022/hash/5f9f9e4da57a94547491a39dc18f1696-Abstract-Conference.html} {Less-forgetting multi-lingual fine-tuning}.
\newblock In \emph{Advances in Neural Information Processing Systems 35: NeurIPS 2022, New Orleans, USA}.

\bibitem[{{Ministry of Digital Transition and Administration Reform}(2024)}]{digitalmorocco2030}
{Ministry of Digital Transition and Administration Reform}. 2024.
\newblock {Digital Morocco 2030: National Strategy}.
\newblock \url{https://www.mmsp.gov.ma/sites/default/files/2024-09/PlaquetteInstitutionnel_18092024_Ang.pdf}.
\newblock Accessed: 2025-05-13.

\bibitem[{NVIDIA(2024)}]{nvidiaA1002024}
NVIDIA. 2024.
\newblock {NVIDIA A100 Tensor Core GPU} datasheet.
\newblock \url{https://www.nvidia.com/content/dam/.../a100-80gb-datasheet.pdf}.
\newblock Accessed May 19, 2025.

\bibitem[{ProtectAI(2025)}]{xlm-lang-detector}
ProtectAI. 2025.
\newblock protectai/xlm-roberta-base-language-detection-onnx.
\newblock \url{https://huggingface.co/protectai/xlm-roberta-base-language-detection-onnx}.
\newblock Accessed May 2025. ONNX version of papluca/xlm-roberta-base-language-detection using the Optimum library. Fine-tuned on the Language Identification dataset.

\bibitem[{Shang et~al.(2025)Shang, Abdine, Khoubrane, Mohamed, Abbahaddou, Ennadir, Momayiz, Ren, Moulines, Nakov, Vazirgiannis, and Xing}]{atlas}
Guokan Shang, Hadi Abdine, Yousef Khoubrane, Amr Mohamed, Yassine Abbahaddou, Sofiane Ennadir, Imane Momayiz, Xuguang Ren, Eric Moulines, Preslav Nakov, Michalis Vazirgiannis, and Eric Xing. 2025.
\newblock \href {https://aclanthology.org/2025.loreslm-1.2/} {Atlas-chat: Adapting large language models for low-resource moroccan arabic dialect}.
\newblock In \emph{Proceedings of the First Workshop on Language Models for Low-Resource Languages}, pages 9--30, Abu Dhabi, United Arab Emirates. Association for Computational Linguistics.

\bibitem[{Strubell et~al.(2019)Strubell, Ganesh, and McCallum}]{strubell-etal-2019-energy}
Emma Strubell, Ananya Ganesh, and Andrew McCallum. 2019.
\newblock \href {https://doi.org/10.18653/v1/P19-1355} {Energy and policy considerations for deep learning in {NLP}}.
\newblock In \emph{Proceedings of the 57th Annual Meeting of the Association for Computational Linguistics}, pages 3645--3650, Florence, Italy. Association for Computational Linguistics.

\bibitem[{Team et~al.(2025)Team, Abbas, Ahmad, Alam, Altinisik, Asgari, Boshmaf, Boughorbel, Chawla, Chowdhury, Dalvi, Darwish, Durrani, Elfeky, Elmagarmid, Eltabakh, Fatehkia, Fragkopoulos, Hasanain, Hawasly, Husaini, Jung, Lucas, Magdy, Messaoud, Mohamed, Mohiuddin, Mousi, Mubarak, Musleh, Naeem, Ouzzani, Popovic, Sadeghi, Sencar, Shinoy, Sinan, Zhang, Ali, Kheir, Ma, and Ruan}]{fanar}
Fanar Team, Ummar Abbas, Mohammad~Shahmeer Ahmad, Firoj Alam, Enes Altinisik, Ehsannedin Asgari, Yazan Boshmaf, Sabri Boughorbel, Sanjay Chawla, Shammur Chowdhury, Fahim Dalvi, Kareem Darwish, Nadir Durrani, Mohamed Elfeky, Ahmed Elmagarmid, Mohamed Eltabakh, Masoomali Fatehkia, Anastasios Fragkopoulos, Maram Hasanain, Majd Hawasly, Mus'ab Husaini, Soon-Gyo Jung, Ji~Kim Lucas, Walid Magdy, Safa Messaoud, Abubakr Mohamed, Tasnim Mohiuddin, Basel Mousi, Hamdy Mubarak, Ahmad Musleh, Zan Naeem, Mourad Ouzzani, Dorde Popovic, Amin Sadeghi, Husrev~Taha Sencar, Mohammed Shinoy, Omar Sinan, Yifan Zhang, Ahmed Ali, Yassine~El Kheir, Xiaosong Ma, and Chaoyi Ruan. 2025.
\newblock \href {http://arxiv.org/abs/2501.13944} {Fanar: An arabic-centric multimodal generative ai platform}.

\bibitem[{Weber et~al.(2024)Weber, Thellmann, Ebert, Flores-Herr, Lehmann, Fromm, and Ali}]{dopolyglots}
Alexander~Arno Weber, Klaudia Thellmann, Jan Ebert, Nicolas Flores-Herr, Jens Lehmann, Michael Fromm, and Mehdi Ali. 2024.
\newblock \href {https://doi.org/10.18653/v1/2024.emnlp-main.1159} {Investigating multilingual instruction-tuning: Do polyglot models demand for multilingual instructions?}
\newblock In \emph{Proceedings of the 2024 Conference on Empirical Methods in Natural Language Processing}, pages 20829--20855, Miami, Florida, USA. Association for Computational Linguistics.

\bibitem[{Wu et~al.(2025)Wu, Zhao, Xu, Liu, and Zong}]{radis}
Junhong Wu, Yang Zhao, Yangyifan Xu, Bing Liu, and Chengqing Zong. 2025.
\newblock \href {https://openreview.net/forum?id=ysAZQ5lCRF} {Boosting {LLM} translation skills without general ability loss via rationale distillation}.
\newblock In \emph{Submitted to ACL Rolling Review - February 2025}.
\newblock Under review.

\bibitem[{Wu et~al.(2024)Wu, Vu, Qu, and Haffari}]{mixture-of-skills}
Minghao Wu, Thuy-Trang Vu, Lizhen Qu, and Gholamreza Haffari. 2024.
\newblock \href {https://aclanthology.org/2024.emnlp-main.787/} {{MIXTURE-OF-SKILLS}: Learning to optimize data usage for fine-tuning large language models}.
\newblock In \emph{Proceedings of the 2024 Conference on Empirical Methods in Natural Language Processing}, pages 14226--14240. Association for Computational Linguistics.

\bibitem[{Xia et~al.(2024)Xia, Yu, Dang, Yang, Wu, Tian, Chang, and Lin}]{rethinking}
Tingyu Xia, Bowen Yu, Kai Dang, An~Yang, Yuan Wu, Yuan Tian, Yi~Chang, and Junyang Lin. 2024.
\newblock \href {http://arxiv.org/abs/2410.09335} {{Rethinking Data Selection at Scale: Random Selection is Almost All You Need}}.
\newblock \emph{arXiv preprint arXiv:2410.09335}.

\bibitem[{Yakhni and Chehab(2025)}]{lebanese}
Silvana Yakhni and Ali Chehab. 2025.
\newblock \href {https://doi.org/10.48550/arXiv.2505.00114} {Fine-tuning llms for low-resource dialect translation: The case of lebanese}.

\bibitem[{Yu et~al.(2024)Yu, Wu, Li, and Yu}]{lions}
Xiao Yu, Qingyang Wu, Yu~Li, and Zhou Yu. 2024.
\newblock \href {https://doi.org/10.18653/v1/2024.emnlp-main.496} {{LIONs}: An empirically optimized approach to align language models}.
\newblock In \emph{Proceedings of the 2024 Conference on Empirical Methods in Natural Language Processing}, pages 8732--8753, Miami, Florida, USA. Association for Computational Linguistics.

\bibitem[{Zellers et~al.(2019)Zellers, Holtzman, Bisk, Farhadi, and Choi}]{zellers2019hellaswag}
Rowan Zellers, Ari Holtzman, Yonatan Bisk, Ali Farhadi, and Yejin Choi. 2019.
\newblock Hellaswag: Can a machine really finish your sentence?
\newblock In \emph{Proceedings of ACL}.

\bibitem[{Zhang et~al.(2024)Zhang, Lee, Fang, Yu, Jia, Jiang, and Barbieri}]{plug}
Zhihan Zhang, Dong-Ho Lee, Yuwei Fang, Wenhao Yu, Mengzhao Jia, Meng Jiang, and Francesco Barbieri. 2024.
\newblock \href {https://doi.org/10.18653/v1/2024.acl-long.379} {{PLUG}: Leveraging pivot language in cross-lingual instruction tuning}.
\newblock In \emph{Proceedings of the 62nd Annual Meeting of the Association for Computational Linguistics (Volume 1: Long Papers)}, pages 7025--7046, Bangkok, Thailand. Association for Computational Linguistics.

\bibitem[{Zhou et~al.(2023)Zhou, Liu, Xu, Iyer, Sun, Mao, Ma, Efrat, Yu, Yu, Zhang, Ghosh, Lewis, Zettlemoyer, and Levy}]{lima}
Chunting Zhou, Pengfei Liu, Puxin Xu, Srini Iyer, Jiao Sun, Yuning Mao, Xuezhe Ma, Avia Efrat, Ping Yu, Lili Yu, Susan Zhang, Gargi Ghosh, Mike Lewis, Luke Zettlemoyer, and Omer Levy. 2023.
\newblock Lima: Less is more for alignment.
\newblock In \emph{Advances in Neural Information Processing Systems (NeurIPS-2023)}, volume~36, pages 55006--55021.

\end{thebibliography}
\bibliographystyle{acl_natbib}

\appendix

\section{Appendix}
\label{sec:appendix}

\subsection{SFT Dataset Preparation}

\subsubsection{Translation Prompt Template}
\label{appendix:translation-prompt}

To ensure consistency and high quality in automatic translation from English to Moroccan Darija, we employed a strict prompt with clear translation and preservation guidelines. The full prompt, as used for dataset generation, is presented below.\\

\noindent \textbf{TRANSLATION\_PROMPT =}

\noindent Translate the following text into Moroccan Darija, written in Arabic script, following these strict instructions:

\noindent DO NOT TRANSLATE the following elements (keep them exactly as they are):

\begin{enumerate}
    \item \textbf{Code blocks}:\\
    Keep \textit{all programming code} (Python, JS, etc.) \textbf{unchanged}.\\
    Translate comments inside code blocks to Darija (Arabic script).
    \item \textbf{Math formulas and LaTeX}:\\
    Preserve all mathematical notation and LaTeX equations. Do not alter them in any way.
    \item \textbf{Technical Terms}:\\
    Keep technical/scientific terms in English unless they have a known Arabic equivalent.\\
    For example:
    \begin{itemize}
        \item Use ``\RL{صيغة}'' for ``formula''
        \item Use ``\RL{معادلة}'' for ``equation''
        \item Use ``\RL{دالة}'' for ``function''
        \item But do not translate terms like ``API'', ``eigenvalue'', ``html'', etc.
    \end{itemize}

    \item \textbf{Proper Names}:\\
    Do not translate names of people, tools, or software (e.g., Python, Newton, GitHub, etc.).
\end{enumerate}

\noindent\textbf{TRANSLATE into Moroccan Darija (Arabic script):}

\begin{itemize}
    \item Regular sentences, explanations, and natural narrative text.
    \item Comments inside code blocks (e.g., \verb|# comment here|).
    \item Use a natural mix of Moroccan Darija and Modern Standard Arabic where appropriate:
    \begin{itemize}
        \item Use Darija for informal or conversational parts.
        \item Use Standard Arabic for formal or widely recognized scientific terms. Do your best to find most suitable translations.
        \item If translating certain phrases would sound awkward, unclear, or artificial in Darija, then translate them in Standard Arabic.
    \end{itemize}
\end{itemize}

\noindent\textbf{GOAL:}

Make the content clear and naturally understandable to Moroccan Arabic speakers while respecting the technical integrity of the original text. Return the translated text without saying anything else.

\noindent\textbf{Examples:}

\noindent \textbf{Example 1 (Code + Explanation)}

\noindent \textbf{Input:}
\begin{quote}
The 'Function' calculate\_mean(data) is used to compute the average of a list of numbers.

\verb|// This function iterates through the list|
\verb|// and calculates the sum.|
Python

\begin{verbatim}
def calculate_mean(data_list):  
    # Calculate sum of elements  
    total_sum = sum(data_list)  
    # Calculate number of elements  
    count = len(data_list)  
    # Return the mean  
    return total_sum / count  
\end{verbatim}
Its 'Complexity' is O(n).
\end{quote}

\noindent \textbf{Output:}
\begin{quote}
\RL{ال }'Function' calculate\_mean(data) \RL{كتستعمل باش نحسبو المعدل ديال شي ليستة ديال الأرقام.}

\verb|//| \RL{هاد الدالة كدور على العناصر}
\verb|//| \RL{وكتحسب المجموع.}
Python

\begin{verbatim}
def calculate_mean(data_list):  
    # Calculate sum of elements  
    total_sum = sum(data_list)  
    # Calculate number of elements  
    count = len(data_list)  
    # Return the mean  
    return total_sum / count  
\end{verbatim}

\RL{ال }'Complexity' \RL{ديالها هي} O(n).
\end{quote}

\textbf{Example 2 (Math Formula)}

\textbf{Input:}
\begin{quote}
Consider the quadratic equation $ax^2 + bx + c = 0$. The solutions can be found using the formula:\\
$x = \frac{-b \pm \sqrt{b^2 - 4ac}}{2a}$. This is known as the quadratic formula.
\end{quote}

\textbf{Output:}
\begin{quote}
\RL{ناخدو المعادلة من الدرجة الثانية} $ax^2 + bx + c = 0$. \RL{الحلول ديالها كتلقاو بهاد الصيغة:}\\
$x = \frac{-b \pm \sqrt{b^2 - 4ac}}{2a}$. \RL{هادي معروفة بصيغة المعادلة من الدرجة الثانية.}
\end{quote}

\textbf{Example 3 (Fully Regular Text)}

\textbf{Input:}
\begin{quote}
This chapter introduces the basic concepts. Please read it carefully.
\end{quote}

\textbf{Output:}
\begin{quote}
\RL{هاد الجزء كيعرف بالمفاهيم الأساسية. عافاك قراه مزيان.}
\end{quote}

\noindent\textbf{Now, translate the following text:}

\subsubsection{Comprehensive Description of Source Datasets}

\textbf{Comprehensive Description of LIMA-1k Dataset:}\\

The LIMA-1k dataset, introduced in \cite{lima}, comprises 1,000 meticulously curated examples specifically designed to efficiently align large language models (LLMs) with minimal computational resources. This dataset originates from an initiative exploring the hypothesis that effective alignment and instruction-tuning do not necessarily require extensive datasets or reinforcement learning from human feedback (RLHF). Instead, the primary goal of LIMA-1k is to demonstrate that alignment can predominantly be achieved by leveraging the knowledge already embedded within pretrained models, with fine-tuning merely instructing the model in stylistic and structural nuances for interacting effectively with users.

LIMA-1k is curated from several high-quality online sources, notably Stack Exchange, wikiHow, and selected Reddit communities (r/WritingPrompts), augmented with manually authored examples to enhance stylistic uniformity and thematic diversity. Table \ref{tab:lima_stats} summarizes the dataset's key statistics.

\begin{table}[H]
\centering
\resizebox{\columnwidth}{!}{%
\begin{tabular}{lccc}
\hline
\textbf{Data Source}   & \textbf{Examples} & \textbf{Avg. Input Length} & \textbf{Avg. Output Length} \\ \hline
Stack Exchange (STEM)             & 200 & 117 & 523   \\
Stack Exchange (Other)            & 200 & 119 & 530   \\
wikiHow                           & 200 & 12  & 1,811 \\
Pushshift Reddit r/WritingPrompts & 150 & 34  & 274   \\
Natural Instructions              & 50  & 236 & 92    \\
Manually Authored (Group A)       & 200 & 40  & 334   \\ \hline
\textbf{Total/Average} & \textbf{1,000}    & \textbf{93}                & \textbf{727}                \\ \hline
\end{tabular}%
}
\caption{Summary statistics of the LIMA-1k dataset \cite{lima}.}
\label{tab:lima_stats}
\end{table}

\noindent Below are representative multi-turn dialogue examples illustrating the conversational coherence and stylistic uniformity achieved by LIMA-1k.

\noindent\textbf{Example 1:}\\
\noindent \textbf{Human:} Tell me an interesting fact about geography. \\

\noindent \textbf{Assistant:} Here are some interesting facts about geography: 
\begin{itemize}
    \item \textbf{Bhutan} is one of the most intriguing places in the world. The Buddhist kingdom is the only carbon-negative country on Earth, with strict environmental laws and cultural norms that preserve its stunning natural landscape. For similar reasons, Bhutan was one of the last countries to introduce television in 1999, and has no traffic lights to this day.
    
    \item There are currently few territories on Earth that are unclaimed by any sovereign state (\textit{terra nullius}). One example is \textbf{Bir Tawil}, along the border between Egypt and Sudan, which was created by a discrepancy between borders drawn in 1899 and 1902. The territory remains unclaimed by either country because it would entail forfeiting the much larger Halaib Triangle, which is claimed by both countries. Bir Tawil has no settled population, but the land is used by Bedouins who roam the area.
    
    \item The \textbf{antipode} of any spot on Earth is the point directly opposite to it on the globe. Interestingly, most land areas’ antipodes fall in the ocean. For instance, the antipode of Spain is in New Zealand, but the antipode of most locations in North America lies in the Pacific Ocean.
\end{itemize}

\noindent\textbf{Example 2:}\\
\noindent \textbf{Human:} I am planning to start a book club with some friends. Can you write an email invitation for the same? \\

\noindent \textbf{Assistant:} Here is an example email invitation for a book club:

\begin{quote}
\textbf{Subject:} Join me for a book club! \\

Hi [Friend’s Name], how are you? \\

I’m starting a book club with some friends, and I’d love to have you join us! \\

The book club is a casual and fun way to read and discuss books that we love or want to learn more about, and also make new friends. We plan to meet more-or-less once a month. \\

We’re thinking of having our first meeting on \textbf{[Date \& Time]}. Could that work for you? \\

Would love to see you there! \\

Thanks, \\
\textbf{[Your Name]}
\end{quote}

The significance of LIMA-1k lies in its exceptional efficiency and effectiveness. Remarkably, despite its modest size of only 750,000 tokens, models fine-tuned on LIMA-1k exhibit robust generalization capabilities and compete closely with extensively trained RLHF models. This dataset convincingly demonstrates the practicality of minimal yet highly curated datasets for achieving high-quality alignment, substantially reducing computational costs and democratizing the capability for effective model alignment.\\

\textbf{Comprehensive Description of DEITA-6k Dataset}\\

DEITA-6k is a meticulously curated dataset of \textbf{6,000 instruction-response pairs} selected from a pool of \textbf{300,000 high-quality public instruction datasets}, including ShareGPT (58k), UltraChat (105k), and WizardLM (143k). It is designed to be the \textit{smallest effective collection} that still achieves state-of-the-art alignment quality after SFT on 7B- and 13B-parameter models.

\paragraph{Selection Methodology} The DEITA pipeline employs three criteria for selection:
\begin{itemize}
    \item \textbf{Complexity (c):} Measured using an EVOL-COMPLEXITY scorer trained to evaluate the reasoning depth of instructions.
    \item \textbf{Quality (q):} Assessed using an EVOL-QUALITY scorer focusing on the helpfulness and informativeness of responses.
    \item \textbf{Diversity:} Ensured by applying a cosine similarity threshold (\(\tau = 0.9\)) over sentence embeddings to avoid redundancy.
\end{itemize}
The final score \( s = c \times q \) ranks the samples, and the highest-ranked examples satisfying diversity constraints form the DEITA-6k dataset.

\textbf{Key Statistics:}

\begin{table}[H]
\centering
\resizebox{\columnwidth}{!}{%
\begin{tabular}{l|l}
\hline
\textbf{Statistic}           & \textbf{Value}                        \\ \hline
Total Examples               & 6,000                                 \\ \hline
Source Pools              & \begin{tabular}[c]{@{}l@{}}ShareGPT (58k), UltraChat (105k), \\ WizardLM (143k)\end{tabular}     \\ \hline
Average Input Length Ceiling & 2,048 tokens                          \\ \hline
Domain Coverage              & Balanced across 8 MT-Bench categories \\ \hline
Complexity Gain vs Random & \begin{tabular}[c]{@{}l@{}}+0.43 MT-Bench (6.46 $\rightarrow$ 6.03\\  on LLaMA-13B)\end{tabular} \\ \hline
Quality Gain vs Random    & \begin{tabular}[c]{@{}l@{}}+3.17\% AlpacaEval \\ (77.08\% $\rightarrow$ 73.91\%)\end{tabular}    \\ \hline
Scaling Insight           & \begin{tabular}[c]{@{}l@{}}3K DEITA items match the performance\\  of 300K examples\end{tabular} \\ \hline
\end{tabular}%
}
\end{table}

\noindent \textbf{Representative Examples:}\\

\noindent \textbf{Example 1: Entain's Competitors Discussion}\\

\noindent \textbf{Human:} list me some of Entain's competitors in their industry\\
\textbf{Assistant:} Entain, formerly known as GVC Holdings, is a multinational sports betting and gaming company. Competitors include:\\
1. Flutter Entertainment\\
2. William Hill\\
3. Bet365\\

\noindent \textbf{Human:} competitors who are also expanding into esports, game development\\
\textbf{Assistant:} Companies include:\\
1. DraftKings\\
2. FanDuel\\
3. Betway\\
4. Unikrn\\

\noindent \textbf{Human:} what studios has Unikrn acquired? Furthermore, elaborate on \"acquired\".\\
\textbf{Assistant:} As of my knowledge cutoff, Unikrn has not acquired any gaming studios. Explanation of acquisitions:\\
1. Tencent/Riot Games: Full acquisition over time.\\
2. Tencent/Supercell: Majority stake through a consortium.\\
3. Activision Blizzard: Complete acquisition of Vivendi Games.\\

\noindent \textbf{Example 2: IP Ownership and Business Model}\\

\noindent \textbf{Human:} I have ownership of the IP of video games created by S2 Games excluding Heroes of Newerth... What is the appeal of this, to any investor?\\
\textbf{Assistant:} Key appeals include:\\
1. Nostalgia and a loyal player base.\\
2. Unique IP and lore for expansion.\\
3. Use of AI-generated art and modern features.\\
4. Platform potential for indie developers.\\
5. Revenue-sharing model incentives.\\
6. Strong growth potential with community engagement.\\

\noindent \textbf{Example 3: Data Warehouse Simplifications}\\

\noindent \textbf{Human:} explain data warehouse architecture in easy language\\
\textbf{Assistant:} A data warehouse collects data from various sources, transforms it, and stores it centrally. Main components:\\
1. Data Sources\\
2. ETL Process\\
3. Data Warehouse\\
4. OLAP System\\
5. Front-End Tools\\

\noindent \textbf{Example 4: Contracted Chief Brand Officer's Role}\\

\noindent \textbf{Human:} Generate a description of a Contracted Chief Brand Officer's role\\
\textbf{Assistant:} A Contracted Chief Brand Officer becomes an integral part of your executive team, leading brand strategy and development, working with internal teams and external creatives to build a competitive brand presence, and providing ongoing strategic advisement.\\

\noindent \textbf{Summary:}\\
\noindent DEITA-6k demonstrates that \textbf{quality, complexity, and diversity can be engineered simultaneously} to produce a highly effective dataset for alignment. Despite its compact size, it enables language models to achieve near-SOTA performance with significantly reduced computational costs.\\

\textbf{Comprehensive Description of TULU-50k Dataset}\\
\subsection*{Comprehensive Description of TULU--50K Dataset}

\textbf{Origin and Purpose:}\\
TULU--50K is a 50\,000-example SFT subset distilled from the public \textsc{T\"ulu~3} SFT pool ($\approx$939 k dialogues) released by the Allen Institute for AI.  
The subset was produced with the \textbf{MIG} (Maximising Information Gain) sampler \citep{mig}, which automatically balances \emph{quality}—measured by DEITA scores—and \emph{diversity} in a semantic label-graph.  
Its primary goal is to offer an open, compact yet high-coverage instruction-following corpus for research on post-training large language models, especially where full-scale SFT mixes are computationally prohibitive.  The data are licensed ODC-BY and are entirely in English.

\noindent \textbf{Selection Method:}\\
MIG first tags every dialogue in the 939 k-sample pool with semantic topic labels, then scores each instance with DEITA quality metrics.  
It models the tag space as a weighted graph, propagates information across semantically-related tags, and greedily selects examples that yield the highest marginal information gain—provably achieving a $(1-1/e)$ approximation of the optimal subset.  
Compared with random or purely quality-based filtering, the resulting 50 k set delivers state-of-the-art downstream scores while cutting sampling time by two orders of magnitude.

\noindent \textbf{Some Examples:}\\
\medskip
\noindent\textbf{Example 1:} \\
\noindent \textbf{Human:} Captain Jake, a somewhat short-tempered commercial airline pilot, has to navigate a complex flight path to avoid airspace controlled by airlines embroiled in corporate squabbles. The flight path consists of a series of vectors in 3-dimensional space. Each vector represents a leg of the journey from one waypoint to another. \\

\begin{itemize}
    \item \textbf{1.} Given the following vectors representing legs of the journey:
    \begin{itemize}
        \item Vector A: \(\mathbf{a} = (3, 2, -1)\)
        \item Vector B: \(\mathbf{b} = (-2, 4, 5)\)
        \item Vector C: \(\mathbf{c} = (1, -3, 2)\)
    \end{itemize}
    Calculate the total displacement vector \(\mathbf{d}\) from the start to the end of the journey. Express \(\mathbf{d}\) in its simplest form.
    
    \item \textbf{2.} Captain Jake needs to maintain a steady altitude change to avoid turbulence caused by corporate squabbles in certain airspace layers. If the journey described by vector \(\mathbf{d}\) must ensure that the vertical displacement (z-component of \(\mathbf{d}\)) is zero, determine the required adjustment vector \(\mathbf{e} = (x, y, z)\) such that the modified displacement vector \(\mathbf{d}' = \mathbf{d} + \mathbf{e}\) satisfies this condition. Additionally, compute the magnitude of the adjustment vector \(\mathbf{e}\).
\end{itemize}

\vspace{0.5em}
\noindent \textbf{Assistant:} Let's break down the problem into the given tasks and solve each part step by step.

\begin{enumerate}
    \item \textbf{Calculate the total displacement vector \(\mathbf{d}\):}
    
    Given vectors:
$$
\begin{aligned}
\mathbf{a} &= (3, 2, -1), \quad \mathbf{b} = (-2, 4, 5), \\
\mathbf{c} &= (1, -3, 2)
\end{aligned}
$$
    
    The total displacement vector is:
    \[
    \mathbf{d} = \mathbf{a} + \mathbf{b} + \mathbf{c}
    \]
    
    Compute each component:
$$
\begin{aligned}
\mathbf{d}_x &= 3 + (-2) + 1 = 2, \\
\mathbf{d}_y &= 2 + 4 + (-3) = 3, \\
\mathbf{d}_z &= -1 + 5 + 2 = 6
\end{aligned}
$$
    
    So, the total displacement vector is:
    \[
    \mathbf{d} = (2, 3, 6)
    \]
    
    \item \textbf{Determine the adjustment vector \(\mathbf{e}\):}
    
    We want \(\mathbf{d}' = \mathbf{d} + \mathbf{e}\) such that the z-component of \(\mathbf{d}'\) is zero:
    \[
    6 + z = 0 \implies z = -6
    \]
    
    Thus, \(\mathbf{e} = (x, y, -6)\). Since only the vertical displacement needs correction, we can choose \(x = 0\) and \(y = 0\) for simplicity. Therefore:
    \[
    \mathbf{e} = (0, 0, -6)
    \]
    
    The magnitude of \(\mathbf{e}\) is:
    \[
    |\mathbf{e}| = \sqrt{0^2 + 0^2 + (-6)^2} = \sqrt{36} = 6
    \]
    
    \textbf{Final Answer:} The displacement vector is \(\mathbf{d} = (2, 3, 6)\), and the magnitude of the adjustment vector \(\mathbf{e}\) is 6.
\end{enumerate}

\bigskip
\noindent\textbf{Example 2:} \\ 
\noindent \textbf{Human:} A dedicated spectator named Alex spends their weekends analyzing basketball player statistics to provide data-driven recommendations. Alex has been tracking the performance of a team throughout the season and has compiled a dataset consisting of the following metrics for each player: points per game (PPG), assists per game (APG), rebounds per game (RPG), and a newly devised metric called \textit{Efficiency Rating} (ER), which is calculated as: 
\[
\text{ER} = (PPG \times 0.4 + APG \times 0.3 + RPG \times 0.3)
\]

\begin{itemize}
    \item \textbf{1.} Alex wants to determine the optimal combination of players in a 5-person lineup to maximize the team's average Efficiency Rating. Given that there are 10 players on the team, develop an algorithm or mathematical model that can identify the best combination. What is the maximum possible average Efficiency Rating for the team?
    
    \item \textbf{2.} After selecting the optimal lineup, Alex decides to project the team's overall performance in the upcoming playoffs. Alex uses a regression model where the team's winning probability in a game is given by:
    \[
    P(\text{win}) = 0.5 + 0.1 \times (\text{average ER of the lineup})
    \]
    Assuming the optimal lineup maximizes the team's average ER to 25, calculate the probability of the team winning a best-of-7 series, where the team must win 4 out of 7 games to advance.
\end{itemize}

\vspace{0.5em}
\noindent \textbf{Assistant:} Let's solve the problem in two parts.

\begin{enumerate}
    \item \textbf{Optimal Lineup to Maximize Average ER:}
    
    \begin{itemize}
        \item Sort the 10 players by their Efficiency Rating (ER).
        \item Select the top 5 players with the highest ER values: 
        \[
        ER_1 \ge ER_2 \ge ER_3 \ge ER_4 \ge ER_5
        \]
        \item Calculate the maximum average ER:
        \[
        \frac{ER_1 + ER_2 + ER_3 + ER_4 + ER_5}{5}
        \]
        \item According to the problem, the maximum average ER is given as 25.
    \end{itemize}
    
    \item \textbf{Projecting the Team's Winning Probability:}
    
    Using the regression model:
    \[
    P(\text{win}) = 0.5 + 0.1 \times 25 = 0.5 + 2.5 = 3.0
    \]
    Since probabilities cannot exceed 1, the model saturates at:
    \[
    P(\text{win}) = 1
    \]
    
    \textit{Probability of Winning a Best-of-7 Series:}
    
    - The team must win 4 out of 7 games. 
    - If \( p = 1 \), the team is certain to win each game.
    - Therefore, the probability of winning the series is effectively:
    \[
    P(\text{series win}) = 1
    \]
    
    \textbf{Final Answer:} The maximum possible average Efficiency Rating is 25, and the team has a theoretical 100\% chance of winning the best-of-7 series under the given model.
\end{enumerate}

\noindent \textbf{Concluding Remarks:}\\
TULU--50K strikes a pragmatic balance between dataset footprint and skill coverage.  
Empirically, models fine-tuned on this 50 k slice match or exceed full-pool baselines on both knowledge-centric (ARC, MMLU) and preference-centric (AlpacaEval v2, MT-Bench) benchmarks while slashing compute and curation time.  
Its transparent provenance, permissive license, and rich per-sample metadata make it an efficient, community-friendly test-bed for exploring data-centric alignment questions and rapid prototyping of instruction-tuned LLMs.

\subsection{Illustrative Examples from Final Darija Datasets}
\label{appendix:translation-examples}

\subsubsection{Illustrative Examples from Darija Lima}
\begin{figure*}[ht]
    \centering
    \includegraphics[width=\textwidth]{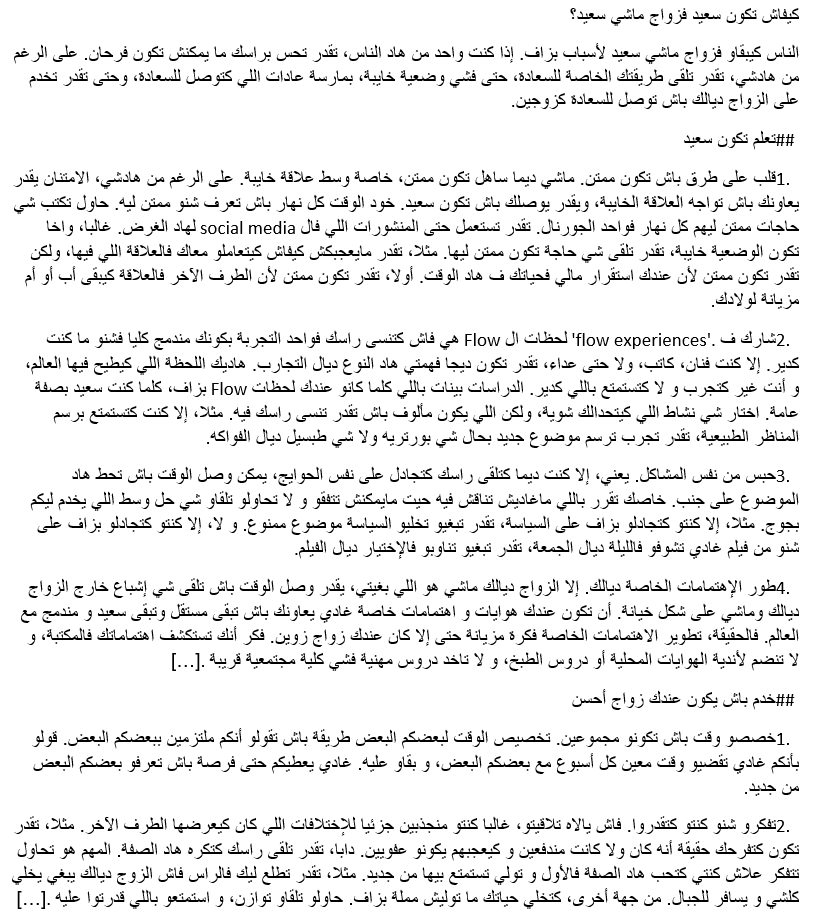}
    \caption{Illustrative example 1 from Darija Lima dataset.}
    \label{fig:lima_exp1}
\end{figure*}

\begin{figure*}[ht]
    \centering
    \includegraphics[width=\textwidth]{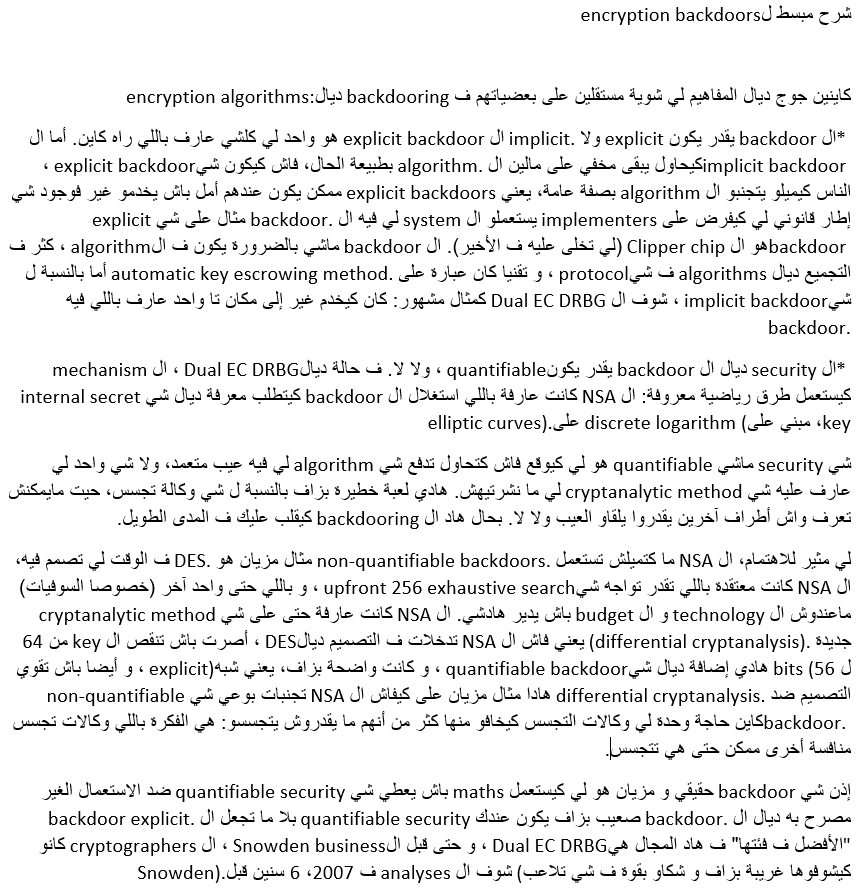}
    \caption{Illustrative example 2 from Darija Lima dataset.}
    \label{fig:lima_exp2}
\end{figure*}

See Figures \ref{fig:lima_exp1} and \ref{fig:lima_exp2}.

\subsubsection{Illustrative Examples from Darija Deita}
\begin{figure*}[ht]
    \centering
    \includegraphics[width=\textwidth]{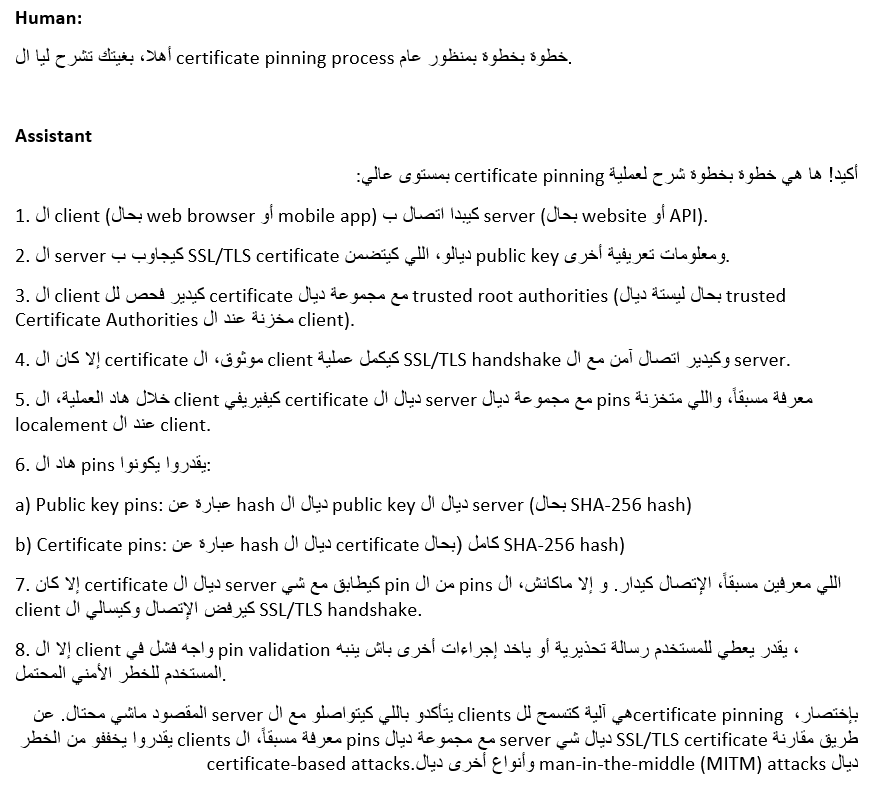}
    \caption{Round 1 of an Illustrative example from Darija Deita dataset.}
    \label{fig:deita_exp1}
\end{figure*}

\begin{figure*}[ht]
    \centering
    \includegraphics[width=\textwidth]{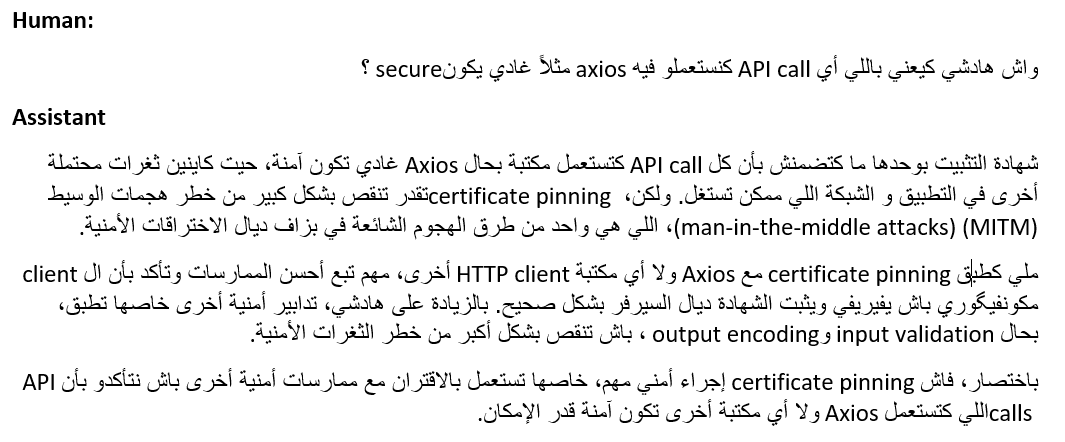}
    \caption{Round 2 of an Illustrative example from Darija Deita dataset.}
    \label{fig:deita_exp2}
\end{figure*}

\begin{figure*}[ht]
    \centering
    \includegraphics[width=\textwidth]{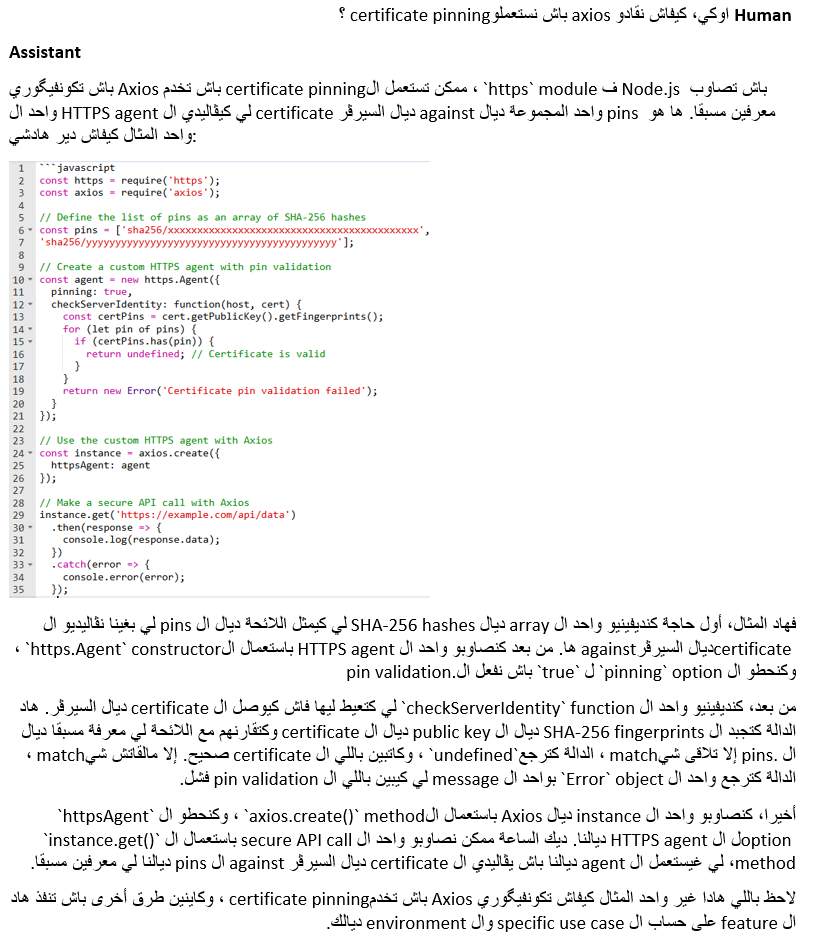}
    \caption{Round 3 of an Illustrative example from Darija Deita dataset.}
    \label{fig:deita_exp3}
\end{figure*}

See Figures \ref{fig:deita_exp1}, \ref{fig:deita_exp2}, and \ref{fig:deita_exp3}.

\subsubsection{Illustrative Examples from Darija TULU}
\begin{figure*}[ht]
    \centering
    \includegraphics[width=\textwidth]{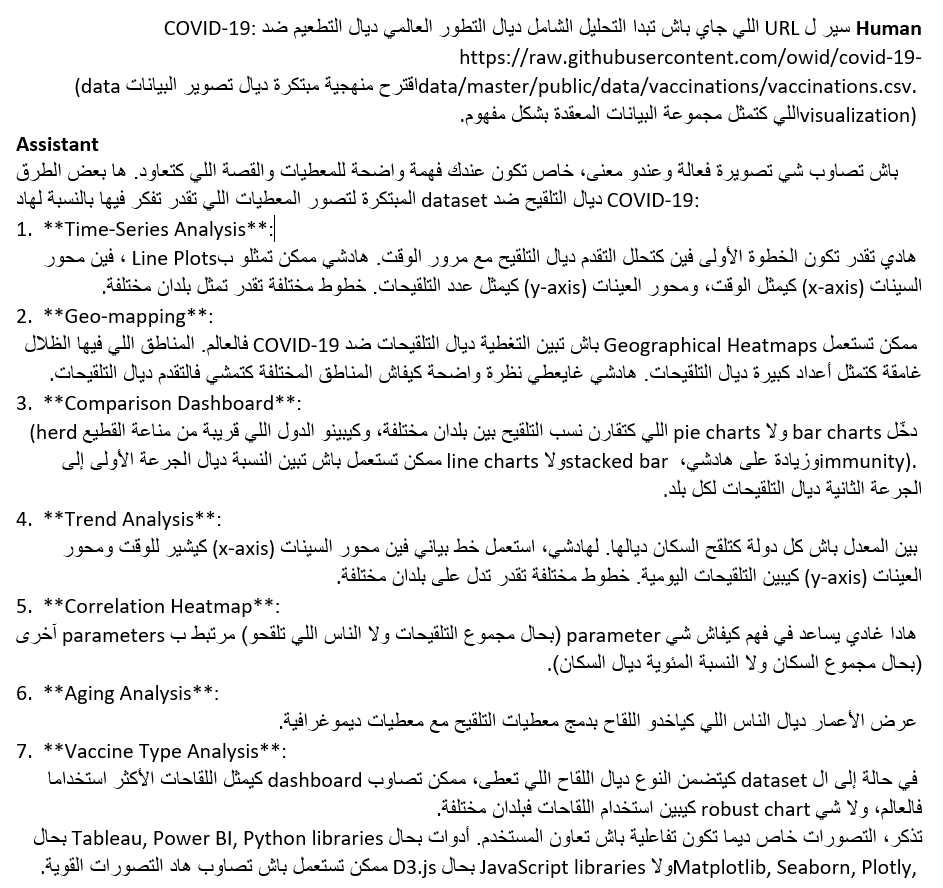}
    \caption{Illustrative example 1 from Darija TULU dataset.}
    \label{fig:tulu_exp1}
\end{figure*}

\begin{figure*}[ht]
    \centering
    \includegraphics[width=\textwidth]{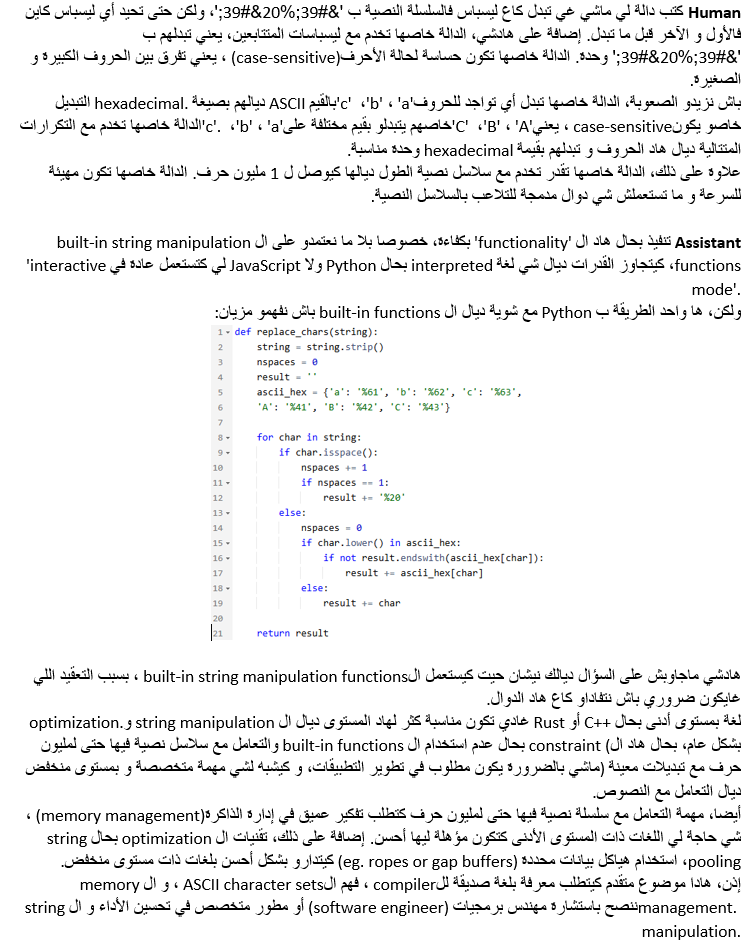}
    \caption{Illustrative example 2 from Darija TULU dataset.}
    \label{fig:tulu_exp2}
\end{figure*}

See Figures \ref{fig:tulu_exp1} and \ref{fig:tulu_exp2}.

\subsection{Efficiency and carbon footprint}
\label{sec:compute-footprint}

\paragraph{Protocol.}
We follow the \textsc{Green\,500} methodology and the recommendations of
\citet{strubell-etal-2019-energy,henderson2020systematic}.  The total
electrical energy \(E\) for a training run is

\[
E \;=\;
\sum_{g\in\{\text{A100},\text{H100}\}}
N_{g}\;t_{g}\;P_{g}\;\text{PUE}\;\eta ,
\]

where \(N_{g}\) and \(t_{g}\) are the number of GPUs and wall–clock
hours, \(P_{g}\) is the average power draw (350 W for A100-80 GB
\citep{nvidia-a100-datasheet}, 500 W for H100-80 GB), \(\text{PUE}=1.3\)
is a modern datacentre power-usage effectiveness, and \(\eta=0.9\)
down-scales for the fact that accelerators seldom sustain their full
TDP for an entire job.

\paragraph{Our training budget.}
Table~\ref{tab:energy-breakdown} lists every experiment reported in
Section~\ref{sec:experiments}.  All ablations on the 4 B backbone
consume a \emph{combined} 10 GPU·h on A100-80 GB, while the
\textsc{Gemmaroc} run uses 48 GPU·h on H100-80 GB.  Plugging these
figures into the formula above gives:

\[
\begin{aligned}
E_{\text{A100}} &= 10 \text{ h}\times 0.35 \text{ kW}\times1.3\times0.9
                 \;=\; 4.1\text{ kWh},\\[2pt]
E_{\text{H100}} &= 48 \text{ h}\times 0.50 \text{ kW}\times1.3\times0.9
                 \;=\; 28.1\text{ kWh},\\[2pt]
E_{\text{total}}&\approx 32\text{ kWh}.
\end{aligned}
\]

With a 2024 global mean grid-carbon intensity of
\(0.40\text{ kg CO}_{2}/\text{kWh}\) \citep{iea2024co2} this implies

\[
C_{\text{ours}}\;\approx\;13 \text{ kg CO}_{2}\text{e}.
\]

\begin{table*}
\centering
\small
\begin{tabular}{@{}lcccc@{}}
\toprule
\textbf{Run} & \textbf{GPUs} & \textbf{Hours} & \textbf{GPU·h} & \textbf{Purpose} \\ \midrule
Gemma-3-4B–LIMA        & 2×A100 & 0.4 & 0.8  & 1 K single-turn \\
Gemma-3-4B–DEITA       & 2×A100 & 0.8 & 1.6  & 6 K dialogue \\
Gemma-3-4B–DEITA$^{+}$ & 2×A100 & 0.8 & 1.6  & +30 \\
Gemma-3-4B–TULU        & 2×A100 & 3 & 6  & 50 K reasoning \\
\midrule
\textit{Subtotal 4 B}  &        &   & \textbf{10} & \\
\midrule
Gemma-3-27B–TULU       & 8×H100 & 6 & 48 & final \textsc{Gemmaroc} \\
\midrule
\textbf{Total}         & —      & — & \textbf{58} & \\ \bottomrule
\end{tabular}
\caption{Revised energy-accounting input reflecting the updated 10 A100 GPU·h for all 4 B experiments.}
\label{tab:energy-breakdown}
\end{table*}

\paragraph{Baselines: Atlas-Chat.}
Atlas-Chat 27 B used LoRA rank 256, 3 epochs over \(500\,\text{k}\) sequences of length 2 048
(\(\simeq 675\,\text{M}\) training tokens)—and used the 
\textsc{Gemma-2-27B} as its base model.  
A recent vLLM benchmark on a single A100-80 GB reports an
\emph{inference} throughput of \(\approx 5\!\times\!10^{2}\) tok s\(^{-1}\) for Gemma-2-27B
\citep{databasemart2024}.  Training carries out a forward
\emph{and} a backward pass; empirical FLOP analyses therefore divide
inference speed by \(\sim\!\!3\).
We conservatively adopt \(r = 55\) tok s\(^{-1}\) GPU\(^{-1}\).

\[
\begin{aligned}
t_{\text{wall}} &= \frac{675 \times 10^{6}}{8\,r\,3600} \;\approx\; 4.26 \times 10^{2} \;\text{h}, \\
t_{\text{GPU}}  &= 8\,t_{\text{wall}} \;\approx\; 3.41 \times 10^{3} \;\text{GPU-h}.
\end{aligned}
\]

Using the NVIDIA A100-80 GB SXM4 average training draw
\(P = 0.35\)\,kW \citep{nvidiaA1002024} and an industry-average
PUE of 1.3 with 90 \% utilisation, the energy and carbon footprints are

\[
\begin{aligned}
E &= t_{\text{GPU}}\,P\,\text{PUE}\,0.9 \;\approx\; 1.4\;\text{MWh}, \\
C &= E \times 0.436\;\text{kg CO\(_2\)e kWh\(^{-1}\)} \;\approx\; 610\;\text{kg CO\(_2\)e}.
\end{aligned}
\]
 \citep{owidCarbon2024}

Finally, at a public on-demand rate of \$2 A100-h\(^{-1}\) the run would
cost \(\sim\$6.8\text{k}\), underscoring the budget advantage of our
own (far cheaper) fine-tune.

\end{document}